\documentclass[10pt,twocolumn,letterpaper]{article}

\usepackage{cvpr}              

\usepackage{graphicx}
\usepackage{amsmath}
\usepackage{amssymb}
\usepackage{booktabs}
\usepackage{wrapfig}

\usepackage{stackengine}
\usepackage{multirow}
\usepackage{tabularx}
\usepackage{wrapfig}
\usepackage{threeparttable}
\usepackage[pagebackref,breaklinks,colorlinks]{hyperref}

\usepackage[capitalize]{cleveref}
\crefname{section}{Sec.}{Secs.}
\Crefname{section}{Section}{Sections}
\Crefname{table}{Table}{Tables}
\crefname{table}{Tab.}{Tabs.}

\def\cvprPaperID{6358} 
\def\confName{CVPR}
\def\confYear{2023}

\begin{document}

\title{OPE-SR: Orthogonal Position Encoding for Designing a Parameter-free Upsampling Module in Arbitrary-scale Image Super-Resolution}

\author{
Gaochao Song$^{1}$ \quad Luo Zhang$^{2}$ \quad Ran Su$^{1}$ \quad Jianfeng Shi$^{3}$
\quad Ying He$^{2}$ \quad Qian Sun$^{3}$ \\
$^{1}$Tianjin University \quad 
$^{2}$Nanyang Technological University \\
$^{3}$Nanjing University of Information Science and Technology \quad
}
\maketitle
\begin{abstract}
   Implicit neural representation (INR) is a popular approach for arbitrary-scale image super-resolution (SR), as a key component of INR, position encoding improves its representation ability. Motivated by position encoding, we propose orthogonal position encoding (OPE) - an extension of position encoding - and an OPE-Upscale module to replace the INR-based upsampling module for arbitrary-scale image super-resolution. Same as INR, our OPE-Upscale Module takes 2D coordinates and latent code as inputs; however it does not require training parameters. This parameter-free feature allows the OPE-Upscale Module to directly perform linear combination operations to reconstruct an image in a continuous manner, achieving an arbitrary-scale image reconstruction.
   As a concise SR framework, our method has high computing efficiency and consumes less memory comparing to the state-of-the-art (SOTA), which has been confirmed by extensive experiments and evaluations. In addition, our method has comparable results with SOTA in arbitrary scale image super-resolution.
Last but not the least, we show that OPE corresponds to a set of orthogonal basis, justifying our design principle. 
\end{abstract}

\section{Introduction}
\label{sec:intro}

In the formation of photograph, the sampling frequency breaks the continuous visual world into discrete pixels of varying precision, while the purpose of the single image super resolution (SISR) task is to restore the original continuous world in the image as much as possible. Therefore, an ideal super resolution task should firstly reconstruct the continuous representation of low-resolution image and then adjust the resolution of target image freely according to actual needs, this is exactly the main idea of arbitrary-scale image super resolution. With the rise of implicit neural representation (INR) in 3D vision \cite{deepsdf,occupancy,convoccu,local,nerf,volume,deepvoxels,texf,srn,pref}, it is possible to represent continuous 3D objects and scenes, which also paves the way for representing continuous image and arbitrary-scale image super-resolution \cite{metasr,liif,xu2021ultrasr,lee2022local}.
\begin{figure}[t]
  \centering
    \includegraphics[scale=0.68]{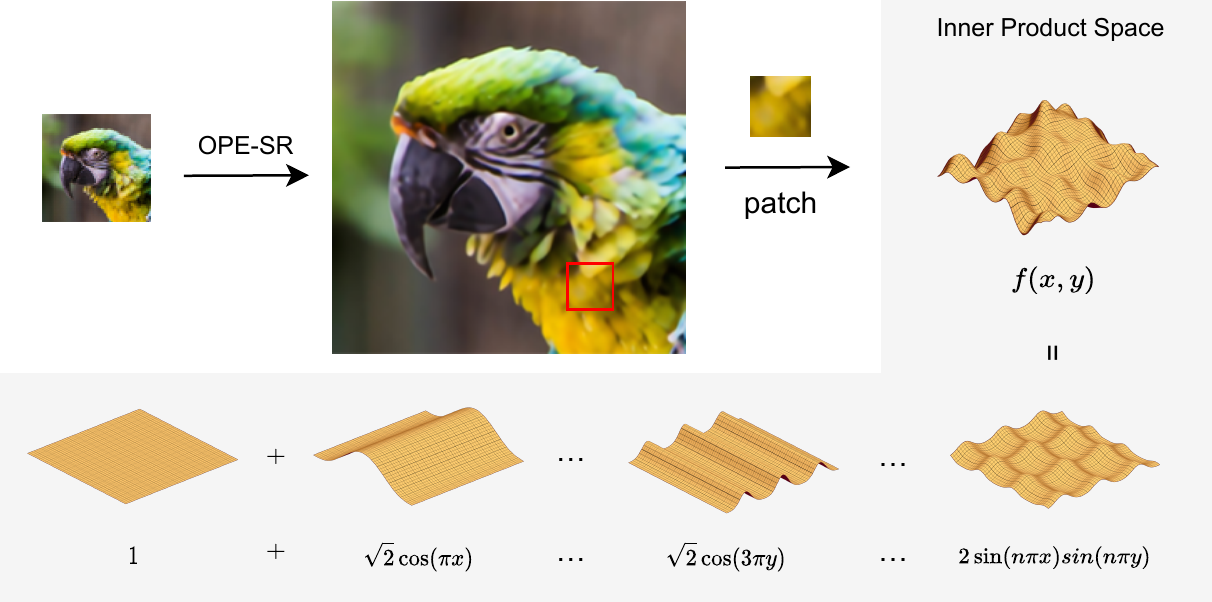}

    \vspace*{-6pt}
    \caption{\textbf{Concept of OPE representation.} A continuous image patch can be decomposed as the linear combination of a group of basic plane wave. OPE-SR: See Fig. \ref{fig:pipline} for our SR framework.}
    \label{concept}
\end{figure}
The existing methods focusing on arbitrary-scale SISR adopt a post-upsampling framework \cite{srsurvey}, in which low-resolution (LR) images first pass through a deep CNN network (encoder) without improving the resolution, and then pass through an INR-based upsampling module together with any specified target resolution to reconstruct high-resolution (HR) images. The INR-based upsampling module establishes a mapping relationship from feature maps (the output of encoder) to target image pixels according to a pre-assigned grid partitioning), and achieves arbitrary-scale with the density of grid in Cartesian coordinate system. To overcome the defect that INR intends to learn low-frequency information, which also known as spectral bias \cite{spectral}, sinusoidal positional encoding is introduced to embed input coordinates to higher dimensions and enables network to learn high-frequency details. This inspires some works of arbitrary-scale SR to improve the representation ability \cite{lee2022local,xu2021ultrasr}.

\begin{figure}[t]
\footnotesize
\centering

\stackunder[2pt]{\includegraphics[width=0.8in]{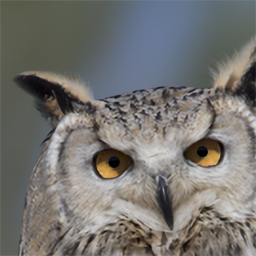}}{LIIF: 32.35}\hspace{-1pt}
\stackunder[2pt]{\includegraphics[width=0.8in]{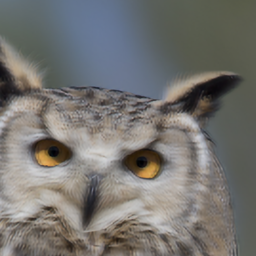}}{LIIF-flip: 28.47}\hspace{1pt}
\stackunder[2pt]{\includegraphics[width=0.8in]{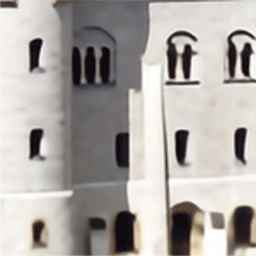}}{LIIF: 25.43}\hspace{-1pt}
\stackunder[2pt]{\includegraphics[width=0.8in]{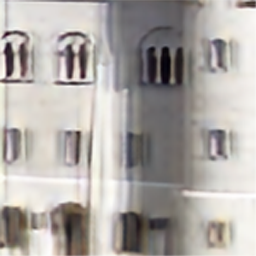}}{LIIF-flip: 23.11}

\stackunder[2pt]{\includegraphics[width=0.8in]{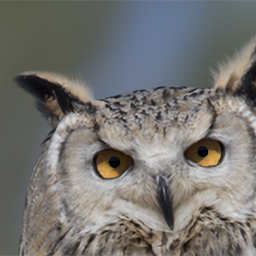}}{OPE: 32.37}\hspace{-1pt}
\stackunder[2pt]{\includegraphics[width=0.8in]{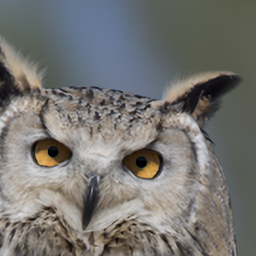}}{OPE-flip: 32.37}\hspace{1pt}
\stackunder[2pt]{\includegraphics[width=0.8in]{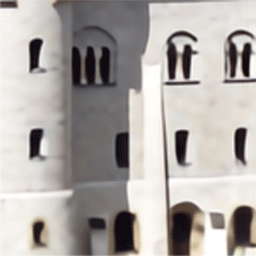}}{OPE: 25.50}\hspace{-1pt}
\stackunder[2pt]{\includegraphics[width=0.8in]{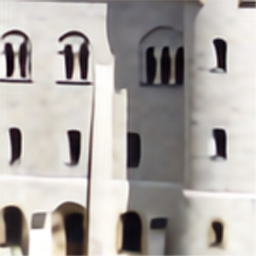}}{OPE-flip: 25.50}

\vspace{-6pt}
\caption{\textbf{Flipping consistency decline (PSNR (dB)).} LIIF-flip: After flipping the input of LIIF \cite{liif} decoder, we get the blurry symmetric output. OPE-flip: Our method get perfect symmetric output. For more samples, see supplementary material.}
\label{liif_flip}
\end{figure}

However, the INR-based upsampling module increases the network complexity since there are two different networks are jointly trained. Besides, as a black-box model, it represents a continuous image with strong dependency on both feature maps and the decoder (e.g. MLP), while it’s representation ability is decreased after flipping the feature map, we call this phenomenon as flipping consistency decline. As shown in Fig. \ref{liif_flip}, after flipping the feature map horizontally before the upsampling module of LIIF, we expect target image has the same flip transformation without other changes, however, the target image appears blurred. It could be that there exists limitation of MLP during learning the symmetry feature of the image.


MLP is a universal function approximator \cite{universal}, which try to fit a mapping function from feature map to the image, therefore, it is reasonable to assume that such process could be solved by an analytical solution.
In this paper, we rethink position encoding from the perspective of orthogonal basis and propose orthogonal position encoding (OPE) for continuous image representation. The linear combination of one-dimensional latent code (extracted vector of feature map over the channel dimension) and OPE can directly reconstruct continuous image patch without using implicit neural function \cite{liif}. To prove OPE's rationality, we analyse it both from functional analysis and 2D-Fourier transform. We further embed it into a parameter-free upsampling module, called OPE-Upscale Module, to replace INR-based upsampling module in deep SR framework, in this case, the deep SR framework can be simplified to the great extent. 


Different from the SOTA work \cite{lee2022local} which enhances MLP by position encoding, we seek for the possibility for building an extending position encoding without MLP. 
By providing a more concise SR framework, our method has high computing efficiency and less memory consumption comparing to SOTA with comparable image performance in arbitrary-scale SR task.




In summary, our contributions are as follows:
\begin{itemize}
    \item We propose a novel position encoding, orthogonal position encoding (OPE), which is in form of 2D-Fourier Series and naturally corresponds to 2D image coordinates. We theoretically prove OPE is correspond to a set of orthogonal basis, which indicates potential representation ability.
    \item The OPE is embeded into our OPE-Upscale Module, which is a parameter-free upsampling module for arbitrary-scale image super-resolution. By providing the more concise SR framework, our method has high computing efficiency and less memory consumption. 
    \item Our OPE-Upscale Module could be easily integrated into existing SR pipeline for interpretable continuous image representation and solves the flipping consistency problem elegantly.
    \item The extensive experiments prove that our method has comparable results with SOTA. Also, our method allows large scale of super-resolution up to $\times$30.
\end{itemize}

\section{Related Work}
\label{sec:relatedworks}

\subsection{Sinusoidal Positional Encoding}
Sinusoidal positional encoding is widely used to erase the negative effects of token order and sequence length in sequence model \cite{vaswani2017attention}, or to guide the image generation as the spatial inductive bias in CNN \cite{lin2021infinitygan,choi2021toward,karras2021alias}. In implicit neural representation, it plays an critical role in solving the limitation of MLP in capturing high-frenquency details, which is referred to as spectral bias \cite{spectral}. By embedding input coordinates into higher dimension, position encoding greatly improves the representation quality in high-frenquency information of implicit 3D scene \cite{nerf,paf} and the follow-up works take it as the default operation to improve representation quality \cite{schwarz2020graf,niemeyer2021giraffe,zhang2020nerf++,liu2020neural}. Inspired by these works, position encoding is preliminarily explored in representating continuous image in arbitrary-scale image SR works \cite{lee2022local,xu2021ultrasr}. Different from the currently used position encoding formulation, our proposed OPE add constant term and take the product of each coordinate embedding as a new term. In the theory part we will reveal this operation aims to construct a more complete orthogonal basis.

\subsection{Orthogonal Basis Representation}
In functional analysis, orthogonal basis decomposes a vector in arbitrary inner product spaces into a group of projections, this concept is widely applied in 2D and 3D tasks. Wavelet transform \cite{daubechies1992ten,mallat1999wavelet} and 2D-Fourier transform are commonly used image analysis methods to clearly separate low-frequency and high-frequency information an image contains. Image moments use two-dimensional orthogonal polynomials \cite{poly1,poly2} to represent image, and are widely used in invariant pattern recognition \cite{pattern1,pattern2}. Image sparse representation inherited this idea of decomposition and performs well in traditional computer vision tasks \cite{sparse1,sparse2,sparse3}. In 3D task, spherical harmonics are an orthogonal basis in space to represent view dependense \cite{sh1,sh2,sh3} and recently is proposed to replace MLP for representing Neural Radiance Fields \cite{plenoxels,nerf}.

\subsection{Deep Learning Based SR Framework}
Based on the upsampling operations and their location in the model, the deep learning based SR can be attributed to four frameworks (see \cite{srsurvey} for a survey): Pre-Upsampling:\cite{srcnn,accurate,memnet,rrn,dr,zero}, Post-Upsampling:\cite{accelerating,srgan,edsr,dsc,rcan}, Progressive-Upsampling:\cite{lap,deeplap,progress} and Iterative Up-and-Down Sampling:\cite{sevenways,dbp,feedback}. For pre-upsampling, the LR image is firstly upsampled by traditional interpolation and then feeded into a deep CNN for reconstructing high-quality details. While it was one of the most popular frameworks with arbitrary-scale factor, it contains side effects like enlarged noise by interpolation and high time and space consumption. The progressive-upsampling and iterative up-and-down sampling frameworks faced with complicated model designing and unclear design criteria. For post-upsampling, the LR image is directly feeded as input of deep CNN, then a trainable upsampling module (e.g. deconvolution \cite{accelerating}, sub-pixel \cite{subpixel} and interpolation convolution \cite{lras}) increases the resolution at the end. Since the huge computational feature extraction process only occurs in low-dimensional space, it has become one of the most mainstream frameworks \cite{lee2022local,swinir,realesrgan}.

\subsection{Arbitrary-scale SR}
Existing arbitrary scale SR works are based on post-upsampling framework. They replace the traditional upsampling module with an INR-based module, a coordinate-based MLP, and shows greatly the convenience and practical potential. \cite{metasr} is the first arbitrary-scale SR work based on CNN, it uses an implicit network to assign an individual convolution kernel for each target pixel to establish the mapping from feature map and dense grid to target image. ArbSR \cite{wang2021learning} puts forward a general plug-in module in a similar way and further solved the scaling problem of different horizontal and vertical scales. SRWarp \cite{son2021srwarp} transforms LR images into HR images with arbitrary shapes via a differential adaptive warping layer. SphereSR \cite{yoon2022spheresr} explores arbitrary-scale on $360^{\circ}$ images for the first time. LIIF \cite{liif} takes coordinates and conditional latent code into MLP and directly predicts target pixel color, it has a intuitive network structure and achieves favorable results on large scale (x6 - x30). LIIF-related follow works focus on the prediction of high-frequency information with position encoding \cite{lee2022local,xu2021ultrasr}.

\begin{figure*}[t]
  \centering
  \includegraphics[width=1.0\textwidth]{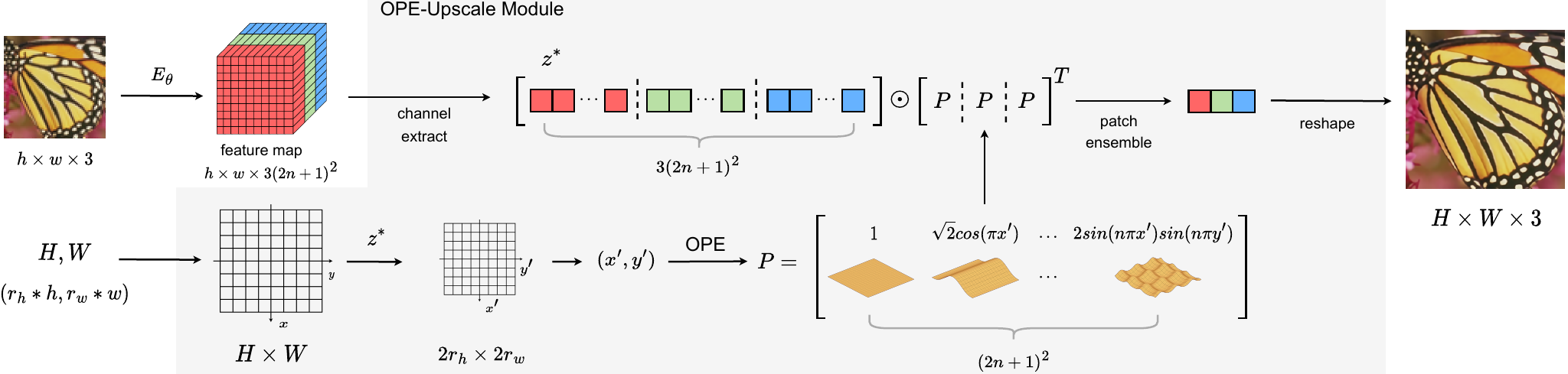}
  \caption{\textbf{OPE-Upscale Module in arbitrary-scale SR framework.} Encoder $E_\theta$ is the only trainable part. With a predefined max frequency $n$ of OPE, the OPE-Upscale Module (grey part) takes feature map from $E_\theta$ and target resolution $H,W$ as input and renders every pixel of target SR image in parallel. $\odot$ is matmul product of $z^* \in \mathbb{R}^{1 \times 3(2n+1)^2}$ and OPEs $\in \mathbb{R}^{3(2n+1)^2 \times 1}$ per RGB channel.}
  \label{fig:pipline}
\end{figure*}
\section{Method}
\label{sec:method}
\subsection{Image Patch Representation with OPE}
In this section we present the formula of continuous image patch representation, which is the basic theory of OPE-Upscale Module. 
The continuous image patch is a continuous binary function $f\in \mathbb{X}$ defined in $[-1, 1]\times[-1, 1]$, where the variables $x, y$ are coordinates in 2D domain mapping from an image pixel, the value $f(x, y)$ is a scalar.
For instance, low resolution image $I_{LR}$ with size $h\times w$, we partition the 2D domain into $h\times w$ grids. For each grid, it represent a low resolution image pixel, also corresponding to a high resolution image patch with size $r_{h}\times r_{w}$, where we expect a high resolution image $I_{SR}$ with size $(H=r_{h}\cdot h , W=r_{w}\cdot w)$. 
For a specific channel (i.e., RGB), the high resolution image patch with size $r_{h}\times r_{w}$, it could be represented as $I\in \mathbb{R}^{r_h \times r_w \times 1}$, each value in $I$,  $I_{(x, y)}$ is treated as a sampling result of a continuous binary function $f\in \mathbb{X}$ defined in $[-1, 1]\times[-1, 1]$, which is estimated as the linear combination of a set of orthogonal basis and projections:
\begin{equation}\label{eq1}
    I_{(x,y)}=f(x,y)\simeq ZP^T
\end{equation}
\begin{equation}\label{eq2}
    P=flat(X^TY)
\end{equation}
\begin{align}\label{eq3}
    X=\gamma(x)&=[1,\sqrt{2}\cos(\pi x),\sqrt{2}\sin(\pi x),\sqrt{2}\cos(2 \pi x),\notag \\
    &...,\sqrt{2}\cos(n \pi x),\sqrt{2}\sin(n \pi x)]\notag \\
    Y=\gamma(y)&=[1,\sqrt{2}\cos(\pi y),\sqrt{2}\sin(\pi y),\sqrt{2}\cos(2 \pi y),\notag \\
    &...,\sqrt{2}\cos(n \pi y),\sqrt{2}\sin(n \pi y)] 
\end{align}
It is worth to note that, in particular, for $x,\ y$, we use central coordinates of the grid to indicate the entire grid. Thus, the pixel value $I_{(x, y)}$ could be calculated for region $[x-1/r_h,x+1/r_h]\times[y-1/r_w,y+1/r_w]$. $\mathbb{X}$ represents the inner product space, for $\forall g,h \in \mathbb{X} $, the inner product is as follows:

\begin{equation}\label{eq5}
    \langle g,h \rangle =\frac{1}{4} \int_{-1}^1\int_{-1}^1 g(x,y)h(x,y) dxdy
\end{equation}
$\gamma(\cdot): \mathbb{R} \rightarrow \mathbb{R}^{1\times (2n+1)}$ represents one variable position encoding with a predefined max frequency $n\in \mathbb{N}$, $flat(\cdot): \mathbb{R}^{(2n+1)\times (2n+1)}\rightarrow \mathbb{R}^{1\times (2n+1)^2}$ represents flattening the 2D matrix to 1D. We call $P\in \mathbb{R}^{1\times (2n+1)^2}$ as orthogonal position encoding (OPE), it performs the linear combination operation with a 1D matrix $Z\in \mathbb{R}^{1\times (2n+1)^2}$ to approximate $f$ rather than MLP as in LIIF \cite{liif}.

\noindent{\bf Theoretical foundation.} This part we analyse our method's rationality from the perspective of functional analysis and fourier analysis. When we take every element $e_{i,\ j}$ of 2D matrix $X^TY$ ($i$-th row,\ $j$-th column) as a binary function, (e.g. $e_{4,5}=2\cos(2\pi x)\sin(2\pi y)$), they satisfy the following relationship:

\begin{equation}\label{eq6}
\langle e_{i_1,j_1},e_{i_2,j_2} \rangle=
\begin{cases}
0, \quad (i_1,j_1)\neq(i_2,j_2)\\
1, \quad (i_1,j_1)=(i_2,j_2)
\end{cases}
\end{equation}
therefore $e_{i,\ j}$ construct a group of orthogonal basis in continuous image space, where OPE contains this and $Z$ is a group of projections on it. At a holistic level, a continuous image patch can be decomposed as a group of basic plane wave (Fig. \ref{concept}).

As for the perspective of fourier analysis, our basis can also be regarded as the real form version after eliminating the complex exponential term of 2D-Fourier basis based on conjugate symmetry when representing real signal. The detailed derivation is provided in supplementary material.

\subsection{OPE-Upscale Module}
In this section, we describe the proposed OPE-Upscale Module. We treat 1D vector latent code as projections on a set of orthogonal basis. OPE with long enough latent code could directly represent a continuous image, however, it suffers from long embedding time and is unstable when representing local high-frequency details, similar to the limitation for fourier transform to describe local information. Considering this issue, we represent continuous image as the seamless stitching of local patches, whose latent codes are extracted from a feature map over the channel dimension. As shown in Fig. \ref{fig:pipline}, the OPE-Upscale module takes both target resolution $H=r_h\cdot h,\ W=r_w\cdot w$ and feature map $\in \mathbb{R}^{h\times w \times 3(2n+1)^2}$ generated from deep encoder $E_\theta$ as inputs, and computes target pixels in parallel. Before this, we need to predefine the max frequency $n$ of OPE and adjust the output channel to $3(2n+1)^2$ in encoder to fit RGB image.

\noindent{\bf Feature map rendering.} 
\begin{figure}[t]
	\centering
    	\includegraphics[width=0.45\textwidth]{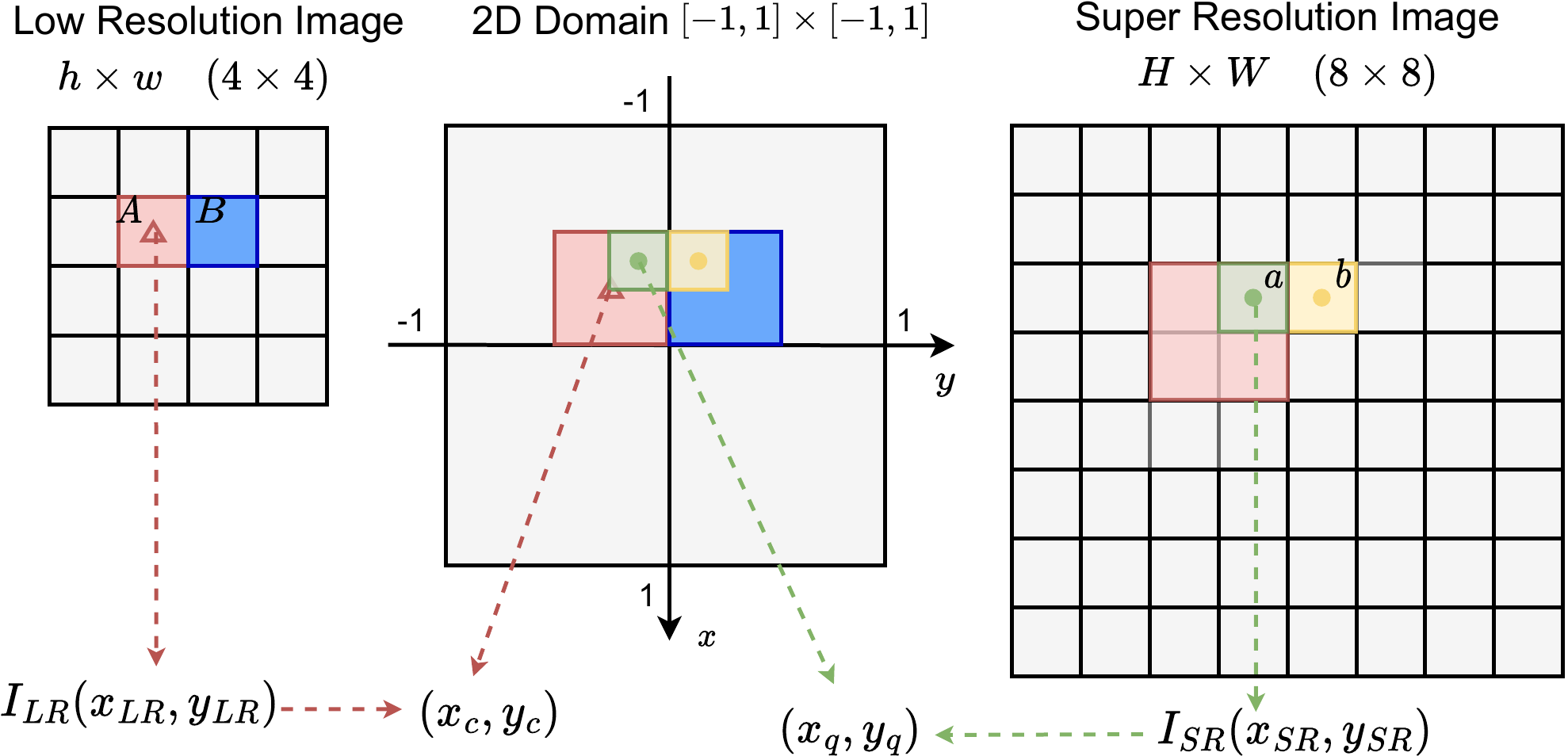}
	\caption{\textbf{Illustration of the mapping relationship from LR image to SR image.} LR image, feature map and SR image are divided in the same domain $[-1,1]\times[-1,1]$. Every LR pixel $I_{LR}(x_{LR},y_{LR})$ corresponds to a latent code with coordinate $(x_c,y_c)$, while every SR pixel $I_{SR}(x_{SR},y_{SR})$ corresponds to $(x_q,y_q)$.}
	\label{fig:fmr}
\end{figure}
As shown in Fig. \ref{fig:fmr}, to render a target image $I_{SR}$ with size $H \times W$ from a low resolution image $I_{LR}$ with size $h \times w$, OPE-Upscale Module firstly divide a 2D domain $[-1,1] \times [-1,1]$ into $H \times W$ regions with equal size, so that every pixel in $I_{SR}$ will be associated with an absolute central coordinates $(x_q, y_q)$ in corresponding region. Secondly, the latent codes in feature map (same dimension with $I_{LR}$) also possess corresponding central coordinates $(x_c, y_c) \in [-1, 1]\times [-1, 1]$ by dividing same 2D domain into $h \times w$ regions, therefore, given a target image pixel with $(x_q, y_q)$, a specific latent code $z^* \in \mathbb{R}^{1 \times 3(2n+1)^2}$ with coordinates $(x_c, y_c)$, which has the smallest distance from $(x_q, y_q)$ could be found. 
As shown in Eq.(\ref{eq7}) and Eq.(\ref{eq8}), a render function $\mathcal{R}$ takes two parts of inputs: $z^{*}$ and $(x_{q}^\prime, y_{q}^\prime)$, to generate final target pixel value as following:

\begin{equation}\label{eq7}
	I_{SR}(x_q,y_q)=\mathcal{R}(z^*,(x_q^\prime,y_q^\prime))
\end{equation}
\begin{equation}\label{eq8}
	x_q^\prime=(x_q-x_c)\cdot h,\quad y_q^\prime=(y_q-y_c)\cdot w
\end{equation}
where $z^*$ is the nearest latent code we found, and $(x_{q}^\prime, y_{q}^\prime)$ are relative coordinates, which are calculated based on Eq.(\ref{eq8}) to rescale the absolute coordinates (in domain $[-1,1] \times [-1,1]$) by times $h$ and $w$, which is taken as input by function $\mathcal{R}$ to render target pixel. $\mathcal{R}$ has the similar calculation as Eq.(\ref{eq1}) while the difference is it repeats OPE 3 times to adapt $z^*$ and calculate linear combination per RGB channel.
In this way, our OPE-Upscale Module successfully deals with arbitrary size $I_{SR}$ by processing each pixel by $\mathcal{R}$, in which feature map rendering process is parameter-free with high computing efficiency and less memory consumption (which has been proved in Sec. \ref{sec:evaluation}).

\noindent{\bf Patch ensemble.} 
There is discontinuity in target image $I_{SR}$ when $(x_q,y_q)$ move from a to b, as shown in Fig. \ref{fig:fmr}, the nearest $z^*$ will change from A to B suddenly. To address this issue, we propose patch ensemble. It contains a local ensemble styled interpolation and the extension of relative coordinate domain. To this end, we extend Eq.(\ref{eq7}) and Eq.(\ref{eq8}) to:
\begin{equation}\label{eq9}
    I_{SR}(x_q,y_q)=\sum_{t\in\{00,01,10,11\}}\frac{s_t}{S}\cdot \mathcal{R}(z^*_t,(x_{q}^\prime,y_{q}^\prime))
\end{equation}
\begin{equation}\label{eq10}
    x_q^\prime=\frac{(x_q-x_t)\cdot h}{2},\quad y_q^\prime=\frac{(y_q-y_t)\cdot w}{2}
\end{equation}
we call Eq.(\ref{eq9}) local ensemble styled interpolation since it takes a similar form of local ensemble in LIIF \cite{liif}.

\begingroup
\setlength{\columnsep}{10pt}
\setlength\intextsep{0pt}
\begin{wrapfigure}{r}{1.5in}
\centering
\includegraphics[height=1.5in]{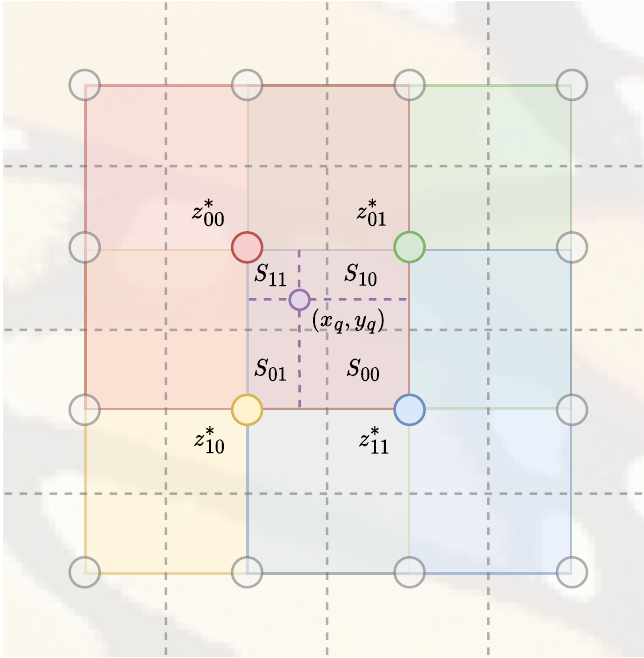}

\vspace{-6pt}
\caption{Patch Ensemble.}
\label{fig:pe}
\end{wrapfigure}
 As shown in Fig. \ref{fig:pe}, instead of finding the nearest one latent code (i.e. $z^*_{00}$), we select the nearest four neighbouring latent codes (i.e. $z^*_{00}, z^*_{01}, z^*_{10}, z^*_{11}$) with corresponding central coordinates $(x_{00}, y_{00}), (x_{01}, y_{01}), \cdots$, we refer $(x_{t}, y_{t})$ in Eq.(\ref{eq10}). Then $x_q^\prime,y_q^\prime$ are calculated based on central coordinates $(x_{t}, y_{t})$ of  $z^*_{00}, z^*_{01}, z^*_{10}$ and $z^*_{11}$ respectively. Eq.(\ref{eq10}) guarantees $x_q^\prime, y_q^\prime$ located in range  $[-1,1]\times[-1,1]$. Eq.(\ref{eq9}) weighted sum of the output of rendering function $\mathcal{R}$, by using the rectangle areas ($s_{11}$, $s_{10}$, $s_{00}$, $s_{01}$) which are finally normalized by $S=\sum_t s_t$, to indicate the contribution of each latent code. To this step, the discontinuity issue in $I_{SR}$ is solved by integrating the adjacent latent codes with different significance and provides a seamless stitching of adjacent patches. To be specific, for four adjacent pixels from low resolution image $I_{LR}$ (i.e. related to $z^*_{00}, z^*_{01}, z^*_{10}$ and $z^*_{11}$), the corresponding patch (the red, green, yellow and blue squares) in super resolution image $I_{SR}$ is not solely depending on the nearest latent code, but considering four neighbouring latent codes with reasonable coefficients.

\endgroup

\subsection{Selection of Max Frequency n}
\label{maxn}
A proper max frequency $n$ (in Eq.(\ref{eq3})) is important since it directly determines the OPE-Upscale Module structure and may have potential effects on different SR scale. 
Given a high resolution image $I_{HR}$ with size $H\times W$, $n$ and $r$, we aim to obtain a feature map with size $\frac{H}{r}\times \frac{W}{r}$, then we re-render the obtained feature map with the selected $n$. By the comparison of the rendered $I_{SR}$ , we present the performance of $n\in\{1, 2, \cdots, 8\}$ under different $r$ values (SR scale), as show in Tab. \ref{represent}, and select the $n$ with the best performance (the details would be discussed Sec. \ref{sec:n}). To be specific, we use Eq.(\ref{eq11}) as the basic theory and use Eq.(\ref{eq12}) to infer the feature map. First, similar to calculate the projection of a normal vector on orthogonal basis, we can calculate projections (or so-called latent code) $Z\in \mathbb{R}^{1\times (2n+1)^2}$ of $f(x,y)$ in Eq.(\ref{eq1}) as follows:

\begin{equation}\label{eq11}
    Z[i]=\frac{1}{4}\int_{-1}^1\int_{-1}^1f(x, y)P[i](x, y)dxdy
\end{equation}
where $P[i](x,y)$ is a binary function taken from the $i$-th position of OPE and $Z[i]$ is the corresponding projection. Based on Eq.(\ref{eq11}) and taking both the discreteness of an image and the design of OPE-Upscale Module into consideration, we calculate the feature map of an image $I_{HR}$ with down-sampling scale $r$ as follows:

\begin{equation}\label{eq12}
    z^*[i]=\frac{1}{4}\sum_{x^\prime}^{2r}\sum_{y^\prime}^{2r}I_{HR}(x^\prime, y^\prime)P[i](x^\prime, y^\prime)
\end{equation}
It can be considered as the inverse operation of Eq.(\ref{eq9}).

\begingroup
\setlength{\columnsep}{10pt}
\setlength\intextsep{0pt}
\begin{wrapfigure}{r}{1.5in}
\centering
\includegraphics[height=1.5in]{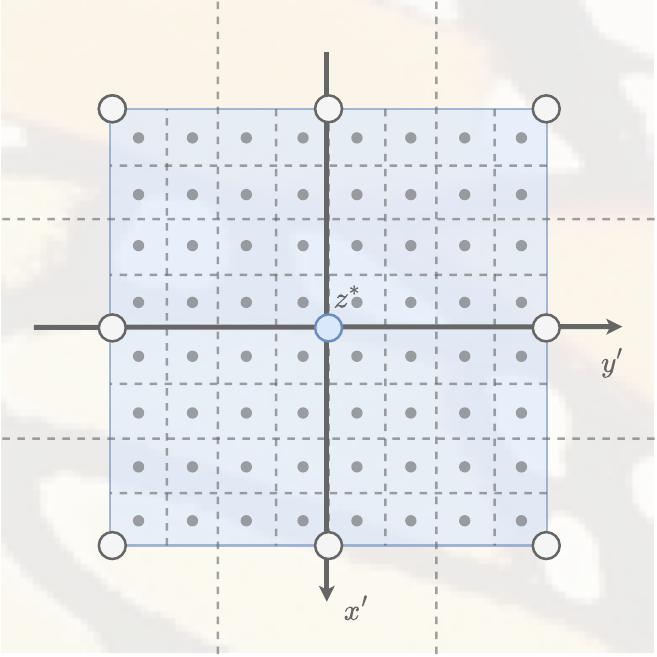}

\vspace{-6pt}
\caption{Inverse operation.}
\label{fig:inverse}
\end{wrapfigure}
 Take Fig. \ref{fig:inverse} as an example. We choose the high resolution image as the ground truth (e.g. HR in Fig. \ref{fig:frequency}), when $r=4$, every latent code $z^*$ corresponds to a $8\times 8$ patch of HR (gray points) in relative coordinate domain (blue region). To calculate the $i$-th position of $z^*$, we multiply every HR pixel value $I_{HR}(x^\prime,y^\prime)$ and basis value $P[i](x^\prime, y^\prime)$ together and finally sum them. After getting the feature map, we render it to the same size of $I_{HR}$ via OPE-Upscale Module and calculate their Peak Signal-to-Noise Ratio (PSNR). We will present experiment result and analysis in Sec. \ref{sec:n}.

\endgroup

\section{Experiments}
\label{sec:exps}
\subsection{Performance of Max Frequency n}
\label{sec:n}
\begin{figure}[t]
\vspace{-6pt}
\footnotesize
\centering

\stackunder[2pt]{\includegraphics[width=0.522in]{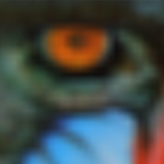}}{n=0}\hspace{-1pt}
\stackunder[2pt]{\includegraphics[width=0.522in]{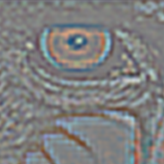}}{n=1}\hspace{-1pt}
\stackunder[2pt]{\includegraphics[width=0.522in]{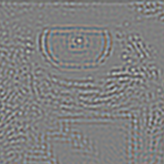}}{n=2}\hspace{-1pt}
\stackunder[2pt]{\includegraphics[width=0.522in]{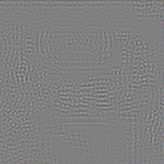}}{n=3}\hspace{-1pt}
\stackunder[2pt]{\includegraphics[width=0.522in]{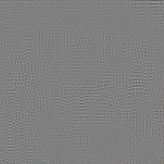}}{n=4}\hspace{-1pt}
\stackunder[2pt]{\includegraphics[width=0.522in]{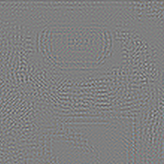}}{n=5}\hspace{-1pt}

\stackunder[2pt]{\includegraphics[width=0.522in]{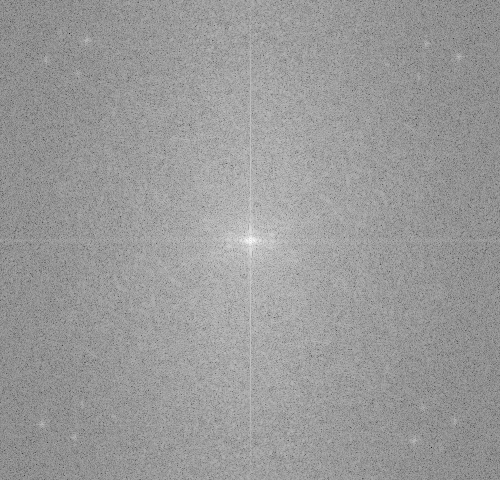}}{HR}\hspace{-1pt}
\stackunder[2pt]{\includegraphics[width=0.522in]{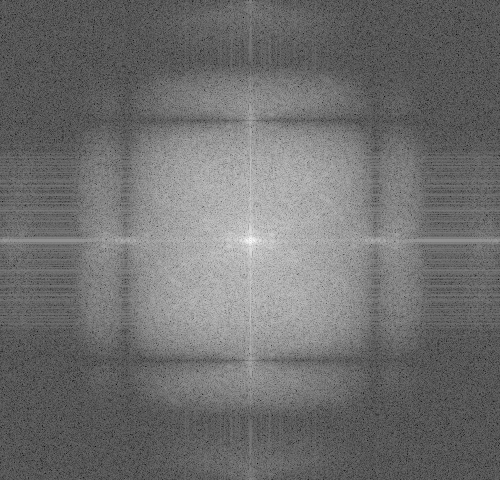}}{n=1}\hspace{-1pt}
\stackunder[2pt]{\includegraphics[width=0.522in]{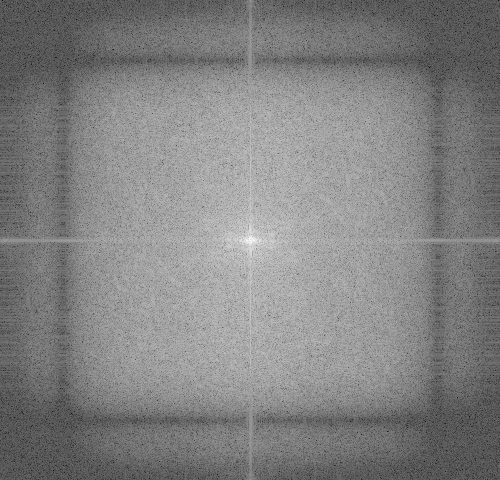}}{n=2}\hspace{-1pt}
\stackunder[2pt]{\includegraphics[width=0.522in]{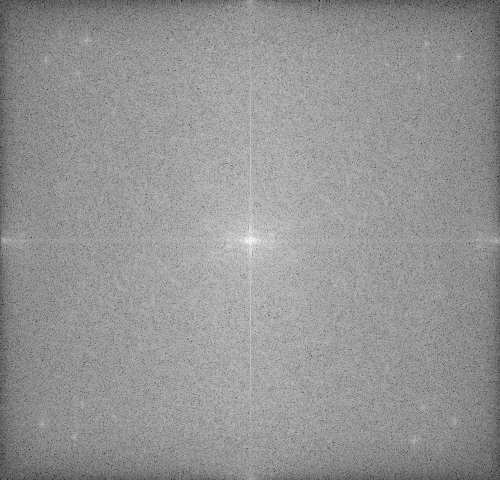}}{n=3}\hspace{-1pt}
\stackunder[2pt]{\includegraphics[width=0.522in]{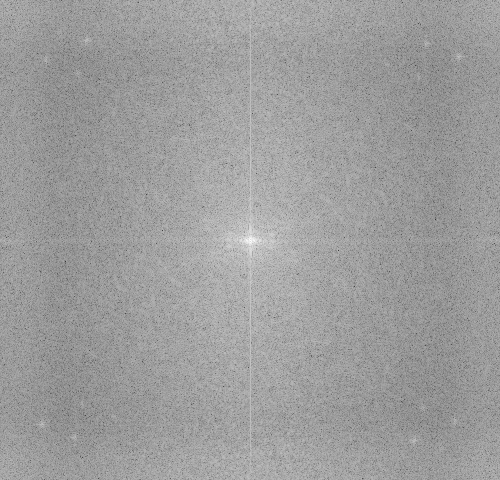}}{n=4}\hspace{-1pt}
\stackunder[2pt]{\includegraphics[width=0.522in]{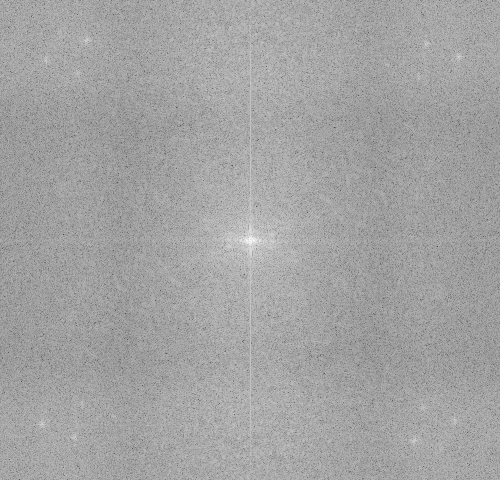}}{n=5}\hspace{-1pt}

\stackunder[2pt]{\includegraphics[width=0.522in]{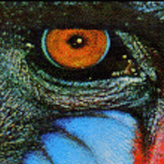}}{HR}\hspace{-1pt}
\stackunder[2pt]{\includegraphics[width=0.522in]{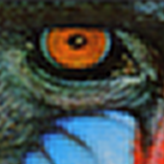}}{21.97}\hspace{-1pt}
\stackunder[2pt]{\includegraphics[width=0.522in]{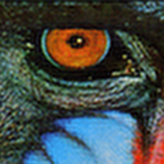}}{25.11}\hspace{-1pt}
\stackunder[2pt]{\includegraphics[width=0.522in]{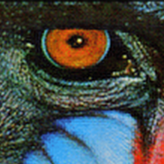}}{\textbf{30.20}}\hspace{-1pt}
\stackunder[2pt]{\includegraphics[width=0.522in]{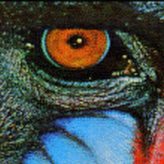}}{29.28}\hspace{-1pt}
\stackunder[2pt]{\includegraphics[width=0.522in]{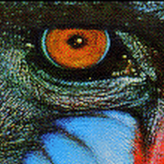}}{22.31}\hspace{-1pt}

\vspace*{-6pt}
\caption{\textbf{Qualitative comparation of different OPE frequency n under scale factor $\times 4$ (PSNR (dB)).} 1-th row: residuals from $n=0$ in image time domain. 2-th row: fourier frequency domain of HR and rendered image with $n$. 3-th row: HR image and rendered image with $n$.}
\label{fig:frequency}
\end{figure}

We sample 50 images from DIV2K validation set \cite{div2k} to explore the representation performance of different $n$ under different scale $r$. We do not use a large $n$ since our method is a local representation. As shown in Tab. \ref{represent} , for a given $r_i$, the optimal sampling frequency is almost always be: $r_i-1$ (the bold number optimum in Tab. \ref{represent}), this phenomenon can also be explained by Nyquist–Shannon sampling theorem. Take $r_i=4$ for example, there are $8\times 8$ sampling points for every latent code to 'fit', hence the max frequency that can be recovered from these sampling points should be less than $4$. Also, we test larger frequency ($n \geq r_i$) until the upper limit $2*r_i$, which is equal to the number of sampling points. We further visualize the effects on reconstructed image for different $n$. As shown in Fig. \ref{fig:frequency}, with scale factor $\times 4$, the larger frequency ($n>3$) will bring redundant high-frequency information and sharpen the image.

This inspires us the selection of max frequency $n$. Since existing arbitrary-scale SR \cite{liif,lee2022local} are training with random scale factor up to $4$, the $n=4-1=3$ already fully covers the ground truth information when training. Obviously larger $n$ would represent more detailed patches of target SR image, however, to avoid redundant high-frequency information, the training scale factor should be larger, which will cause more time and memory consumption when training. For these consideration, we finally choose $n=3$ as our OPE-Upscale Module.

\begin{table}[t]
    \centering
    \renewcommand\arraystretch{1.2}
    \scalebox{0.7}{
\begin{tabular}{c|ccccccc} 
n &$\times 2$ & $\times 3$ &$\times 4$ &$\times 5$ &$\times 6$ &$\times 7$ & $\times 8$\\
\hline
1 &\textbf{31.1951}	&28.6083	&26.4424	&25.0485	&24.1114	&23.3423	&22.7898\\
2 &30.7472	&\textbf{33.6586}	&31.2091	&28.8022	&27.3701	&26.1913	&25.3838\\
3 &22.1871	&33.6585	&\textbf{35.1983}	&32.4011	&30.6135	&28.8964	&27.8159\\
4 &12.1230	&28.6083	&34.9631	&\textbf{34.6294}	&34.0979	&31.4462	&30.2865\\
5 &- 	    &22.8465 	&29.9512	&34.6293	&\textbf{37.3704}	&33.7285 	&32.8190\\
6 &-	    &22.8465 	&24.3122	&32.4011 	&37.1506 	&\textbf{35.3046}	&35.8250\\
7 &-	    &-	        &19.1593	&28.8022 	&33.3039	&35.3046            &\textbf{39.1160}\\
8 &-	    &-	        &12.0863	&25.0485	&29.0966	&33.7286 	&38.9863 
\\
\end{tabular}
}
\vspace*{-6pt}
\caption{\textbf{Representation performance (PSNR (dB)).} The best value for each upsampling factor is bolded.}
\label{represent}
\end{table}

\begin{figure*}[t]
\footnotesize
\centering

\raisebox{0.2in}{\rotatebox{90}{$\times8$ scale}}
\includegraphics[width=1.392in, height=1.044in]{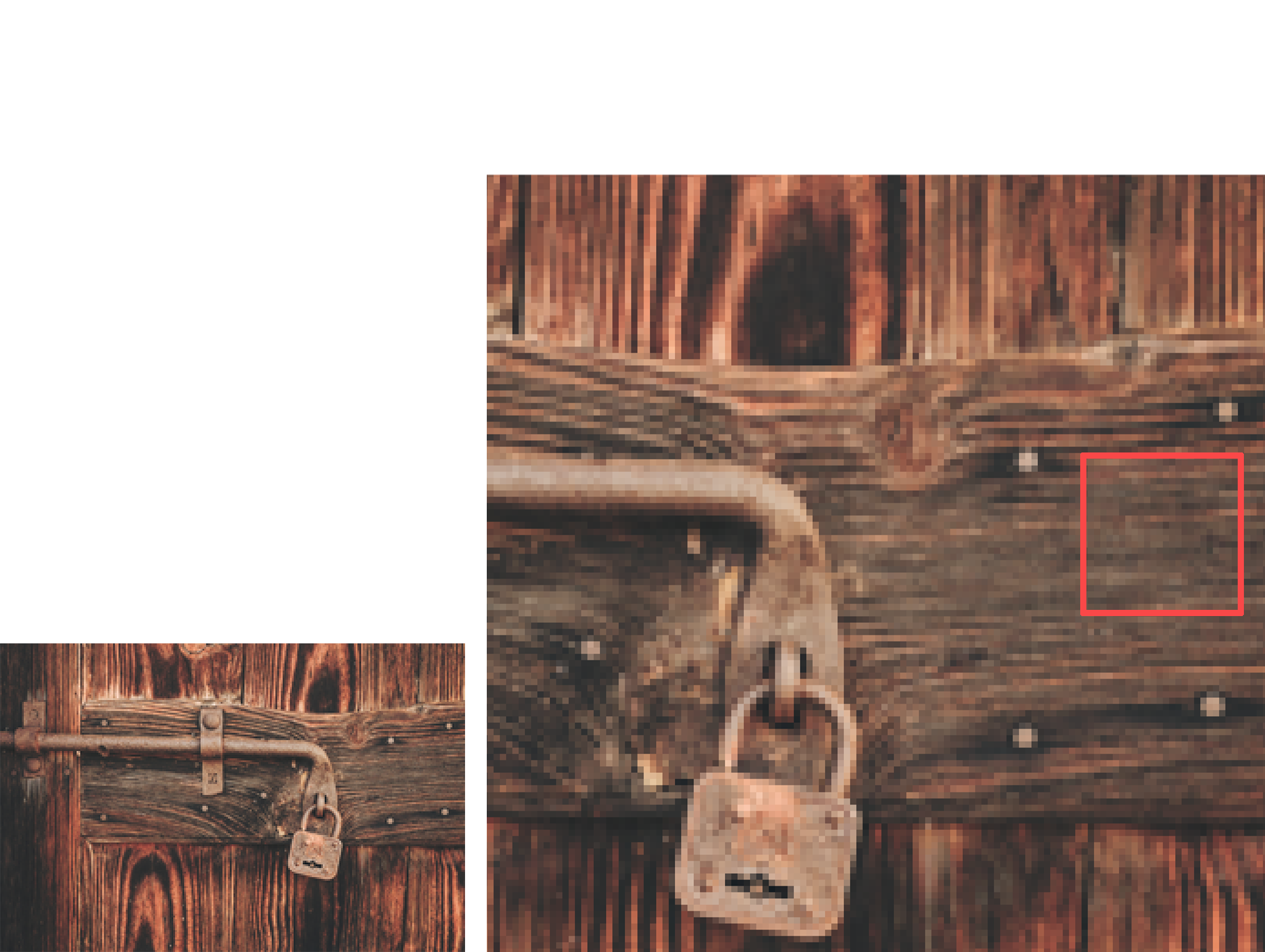}\hspace{-1pt}
\includegraphics[width=1.044in]{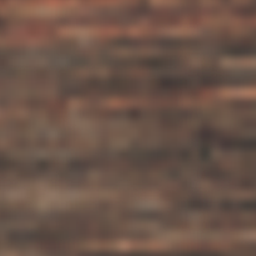}\hspace{-1pt}
\includegraphics[width=1.044in]{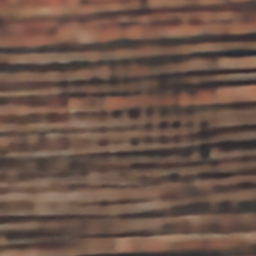}\hspace{-1pt}
\includegraphics[width=1.044in]{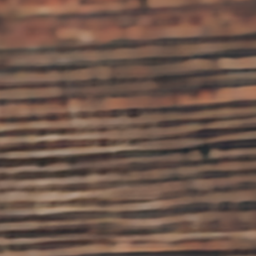}\hspace{-1pt}
\includegraphics[width=1.044in]{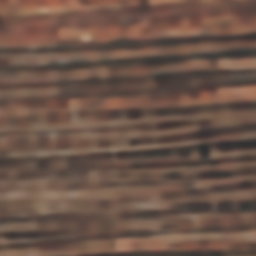}\hspace{-1pt}
\includegraphics[width=1.044in]{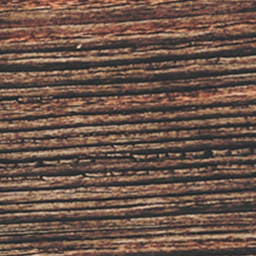}\hspace{-1pt}

\raisebox{0.2in}{\rotatebox{90}{$\times12$ scale}}
\stackunder[2pt]{\includegraphics[width=1.392in, height=1.044in]{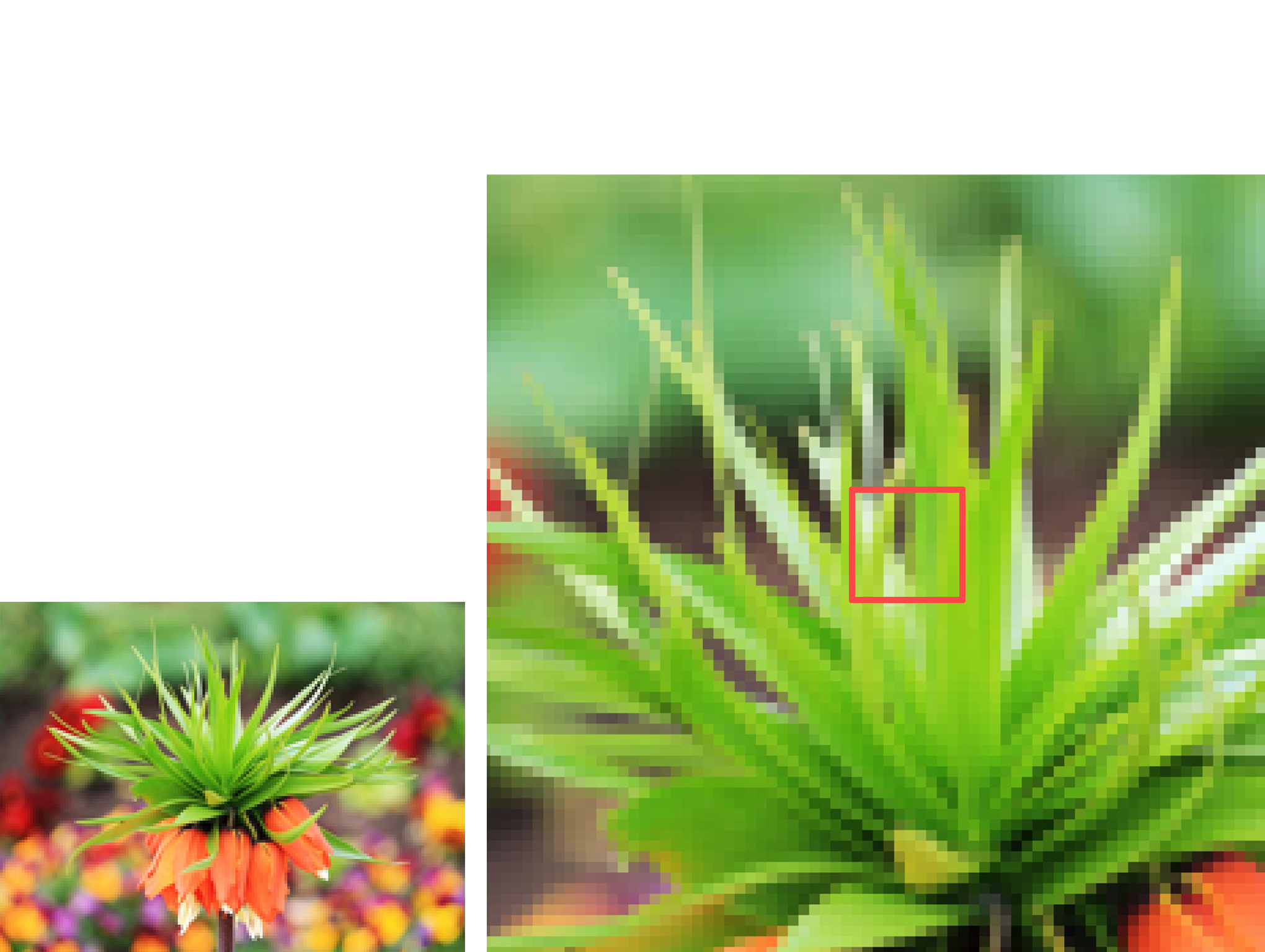}}{LR Image}\hspace{-1pt}
\stackunder[2pt]{\includegraphics[width=1.044in]{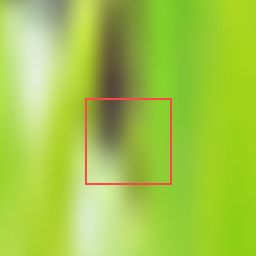}}{Bicubic}\hspace{-1pt}
\stackunder[2pt]{\includegraphics[width=1.044in]{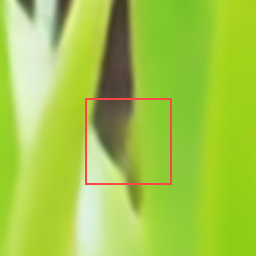}}{LIIF \cite{liif}}\hspace{-1pt}
\stackunder[2pt]{\includegraphics[width=1.044in]{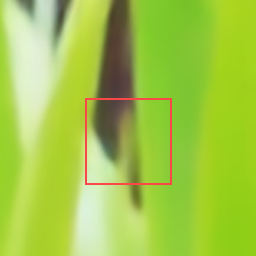}}{LTE \cite{lee2022local}}\hspace{-1pt}
\stackunder[2pt]{\includegraphics[width=1.044in]{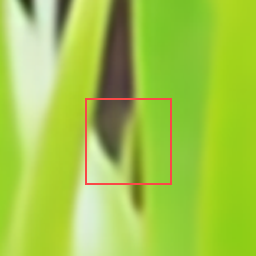}}{\textbf{OPE-SR} (ours)}\hspace{-1pt}
\stackunder[2pt]{\includegraphics[width=1.044in]{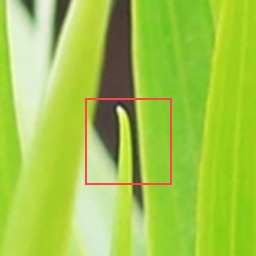}}{GT}\hspace{-1pt}

\vspace*{-6pt}
\caption{\textbf{Qualitative comparison} with SOTA methods for arbitrary-scale SR. RDN \cite{rdn} is used as encoder for all methods.}
\label{results_img}
\end{figure*}

\begin{table*}[t]
\centering
\scalebox{0.75}{
\begin{tabular}{c|ccc|ccccc}
\multirow{2}{*}{Method} & \multicolumn{3}{c|}{In-scale} & \multicolumn{5}{c}{Out-scale} \\
 & $\times2$ & $\times3$ & $\times4$ & $\times6$ & $\times12$ & $\times18$ & $\times24$ & $\times30$\\
\hline
Bicubic \cite{edsr} & 31.01 & 28.22 & 26.66 & 24.82  & 22.27 & 21.00 & 20.19 & 19.59 \\
\hline
EDSR-baseline \cite{edsr} & 34.55 & 30.90 & 28.94 & - & - & - & - & - \\
EDSR-baseline-MetaSR$^\sharp$ \cite{metasr,liif} & 34.64 & 30.93 & 28.92 & 26.61 & 23.55 & 22.03 & 21.06 & 20.37 \\
EDSR-baseline-LIIF \cite{liif} & 34.67 / 1702 & 30.96 / 1277 & 29.00 / 1144 & 26.75 / 1046 & 23.71 / 965 & 22.17 / 953 & 21.18 / 951 & 20.48 / 947 \\
EDSR-baseline-LTE \cite{lee2022local} & 34.72 / 1158 & 31.02 / 1079 & 29.04 / 1045 & 26.81 / 1023 & 23.78 / 1007 & 22.23 / 1005 & 21.24 / 1003 & 20.53 / 1000\\
EDSR-baseline-OPE (ours) & 34.34 / \textcolor[rgb]{1,0,0}{476} & \textcolor[rgb]{0,0,1}{30.94} / \textcolor[rgb]{1,0,0}{395} & \textcolor[rgb]{0,0,1}{29.02} / \textcolor[rgb]{1,0,0}{364} & \textcolor[rgb]{0,0,1}{26.77} / \textcolor[rgb]{1,0,0}{348} & \textcolor[rgb]{0,0,1}{23.74} / \textcolor[rgb]{1,0,0}{322} & \textcolor[rgb]{0,0,1}{22.21} / \textcolor[rgb]{1,0,0}{318} & \textcolor[rgb]{0,0,1}{21.21} / \textcolor[rgb]{1,0,0}{314}& \textcolor[rgb]{0,0,1}{20.52} / \textcolor[rgb]{1,0,0}{311}\\
\hline
RDN-MetaSR$^\sharp$ \cite{metasr,liif} & 35.00 & 31.27 & 29.25 & 26.88 & 23.73 & 22.18 & 21.17 & 20.47 \\
RDN-LIIF \cite{liif} & 34.99 / 3107 & 31.26 / 2073 & 29.27 / 1513 & 26.99 / 1248 & 23.89 / 1025 & 22.34 / 994 & 21.31 / 991 & 20.59 / 972\\
RDN-LTE \cite{lee2022local} & 35.04 / 2549 & 31.32 / 1839 & 29.33 / 1420 & 27.04 / 1184 & 23.95 / 1049 &22.40 / 1027 & 21.36  / 1025& 20.64 / 1014 \\
RDN-OPE (ours) & 34.52 / \textcolor[rgb]{1,0,0}{2277} & 31.17  / \textcolor[rgb]{1,0,0}{1497} & \textcolor[rgb]{0,0,1}{29.26} / \textcolor[rgb]{1,0,0}{1039} & \textcolor[rgb]{0,0,1}{26.98} / \textcolor[rgb]{1,0,0}{813} & \textcolor[rgb]{0,0,1}{23.91} / \textcolor[rgb]{1,0,0}{663} & \textcolor[rgb]{0,0,1}{22.36} / \textcolor[rgb]{1,0,0}{623} & \textcolor[rgb]{0,0,1}{21.34} / \textcolor[rgb]{1,0,0}{596} & \textcolor[rgb]{0,0,1}{20.63} / \textcolor[rgb]{1,0,0}{590} \\
\end{tabular}
}
\vspace*{-6pt}
\caption{\textbf{Quantitative comparison} with SOTA methods for arbitrary-scale image SR on DIV2K validation set (PSNR (dB) / process time (ms/Img)). $^\sharp$ indicates implementation in LIIF \cite{liif}. With a parameter-free upsampling module, we narrow the gap between SOTA and ours in most results less than 0.1dB (\textcolor[rgb]{0,0,1}{blue} number) and obtain the shortest inference time (\textcolor[rgb]{1,0,0}{red} number).}
\label{div2k_val}
\end{table*}
\begin{table*}[t]
\centering
\renewcommand\arraystretch{1.2}
\setlength{\tabcolsep}{1.2pt}
\scriptsize{
\begin{tabular}{c
|>{\centering\arraybackslash}p{0.67cm}>{\centering\arraybackslash}p{0.67cm}>{\centering\arraybackslash}p{0.67cm}
|>{\centering\arraybackslash}p{0.67cm}>{\centering\arraybackslash}p{0.67cm}
|>{\centering\arraybackslash}p{0.67cm}>{\centering\arraybackslash}p{0.67cm}>{\centering\arraybackslash}p{0.67cm}
|>{\centering\arraybackslash}p{0.67cm}>{\centering\arraybackslash}p{0.67cm}
|>{\centering\arraybackslash}p{0.67cm}>{\centering\arraybackslash}p{0.67cm}>{\centering\arraybackslash}p{0.67cm}
|>{\centering\arraybackslash}p{0.67cm}>{\centering\arraybackslash}p{0.67cm}
|>{\centering\arraybackslash}p{0.67cm}>{\centering\arraybackslash}p{0.67cm}>{\centering\arraybackslash}p{0.67cm}
|>{\centering\arraybackslash}p{0.67cm}>{\centering\arraybackslash}p{0.67cm}}
\multirow{3}{*}{Method} & \multicolumn{5}{c|}{Set5} & \multicolumn{5}{c|}{Set14}
& \multicolumn{5}{c|}{B100} & \multicolumn{5}{c}{Urban100} \\
\cline{2-21}
& \multicolumn{3}{c|}{In-scale} & \multicolumn{2}{c|}{Out-scale}
& \multicolumn{3}{c|}{In-scale} & \multicolumn{2}{c|}{Out-scale}
& \multicolumn{3}{c|}{In-scale} & \multicolumn{2}{c|}{Out-scale}
& \multicolumn{3}{c|}{In-scale} & \multicolumn{2}{c}{Out-scale} \\
& $\times2$ & $\times3$ & $\times4$ & $\times6$ & $\times8$
& $\times2$ & $\times3$ & $\times4$ & $\times6$ & $\times8$
& $\times2$ & $\times3$ & $\times4$ & $\times6$ & $\times8$
& $\times2$ & $\times3$ & $\times4$ & $\times6$ & $\times8$\\
\hline
RDN \cite{rdn} & 38.24 & 34.71 & 32.47 & - & -
& 34.01 & 30.57 & 28.81 & - & -
& 32.34 & 29.26 & 27.72 & - & -
& 32.89 & 28.80 & 26.61 & - & - \\
RDN-MetaSR$^\sharp$ \cite{liif, metasr} & 38.22 & 34.63 & 32.38 & 29.04 & 26.96
& 33.98 & 30.54 & 28.78 & 26.51 & 24.97
& 32.33 & 29.26 & 27.71 & 25.90 & 24.83
& 32.92 & 28.82 & 26.55 & 23.99 & 22.59 \\
RDN-LIIF \cite{liif} & 38.17 & 34.68 & 32.50 & 29.15 & 27.14
& 33.97 & 30.53 & 28.80 & 26.64 & 25.15
& 32.32 & 29.26 & 27.74 & 25.98 & 24.91
& 32.87 & 28.82 & 26.68 & 24.20 & 22.79 \\
RDN-LTE \cite{lee2022local} & 38.23 & 34.72 & 32.61 & 29.32 & 27.26
& 34.09 & 30.58 & 28.88 & 26.71 & 25.16
& 32.36 & 29.30 & 27.77 & 26.01 & 24.95
& 33.04 & 28.97 & 26.81 & 24.28 & 22.88 \\
RDN-OPE (ours) & 37.60 & \textcolor[rgb]{0,0,1}{34.59} & \textcolor[rgb]{0,0,1}{32.47} & \textcolor[rgb]{0,0,1}{29.17} & \textcolor[rgb]{0,0,1}{27.22}
& 33.39 & \textcolor[rgb]{0,0,1}{30.49} & \textcolor[rgb]{0,0,1}{28.80} & \textcolor[rgb]{0,0,1}{26.65} & \textcolor[rgb]{0,0,1}{25.17}
& 32.05 & \textcolor[rgb]{0,0,1}{29.19} & \textcolor[rgb]{0,0,1}{27.72} & \textcolor[rgb]{0,0,1}{25.96} & \textcolor[rgb]{0,0,1}{24.91}
& 31.78 & 28.63   & 26.53  & 24.06      & 22.70 \\
\end{tabular}
}
\vspace*{-6pt}
\caption{\textbf{Quantitative comparison} with SOTA methods for arbitrary-scale image SR on benchmark datasets (PSNR (dB)). $^\sharp$ indicates implementation in LIIF \cite{liif}. We narrow the gap between SOTA and ours in most results less than 0.15dB (\textcolor[rgb]{0,0,1}{blue} number). For large scale factor, we keep comparable results to MetaSR \cite{metasr} and LIIF \cite{liif}. The defect in low scale factor will be analysed in Sec. \ref{sec:discuss}.}
\label{tab:Quan_Bench}
\vspace{-10pt}
\end{table*}

\subsection{Training}
\noindent{\bf Datasets.}
Similar to \cite{liif,lee2022local}, we use DIV2K dataset \cite{div2k} of NTIRE 2017 Challenge \cite{ntire} for training.
For testing, we use DIV2K validation set \cite{div2k} with 100 images and four benchmark datasets: Set5 \cite{set5}, Set14 \cite{set14}, B100 \cite{b100}, and Urban100 \cite{urban100}. We use PSNR as evaluation measurement.

\noindent{\bf Implementation details.} 
We mainly follow the prior implementation \cite{liif,lee2022local} for arbitrary-scale SR training after replacing their upsampling module with OPE-Upscale. We use EDSR-baseline \cite{edsr} and RDN \cite{rdn} without their upsampling modules as the encoder, which is the only trainable part of our network. We use $48\times 48$ patches as inputs, L1 loss and Adam \cite{adam} optimizer for optimization. For the arbitrary-scale down-sampling method, we use bicubic resizing in Pytorch \cite{pytorch}. The network was trained for 1000 epochs with batch size 16, while the initial learning rate is 1e-4 and decayed by factor 0.5 every 200 epochs. 



\subsection{Evaluation}
\label{sec:evaluation}

\noindent{\bf Quantitative results.} 
Tab. \ref{div2k_val} and Tab. \ref{tab:Quan_Bench} report quantitative results of OPE and the SOTA arbitrary-scale SR methods on the DIV2K validation set and the benchmark datasets. It is worth noting that we focus on finding an alternative of MLP with position encoding, rather than enhancing it like LTE \cite{lee2022local}. We observe that our method achieves comparable results (less than 0.1dB on DIV2K and less than 0.15dB on benchmark), which indicates that our method is a feasible  analytical solution with good performance and efficient parameter-free module.
As shown in Tab. \ref{div2k_val}, EDSR \cite{edsr} and RDN \cite{rdn} are our selected encoders, and our method achieves the highest efficiency (i.e. the shortest inference time in red number) comparing to all the other baselines with both encoders. The higher the scale factor, the better result we achieve. Specifically, in out-scale SR ($\times 6$ to $\times 30$), our method outperforms most baselines and just has a small gap with LTE (less than 0.1dB). Such results demonstrate that our method has rich representation capability. We also compared with the benchmark dataset, as shown in Tab. \ref{tab:Quan_Bench}, we keep comparable results to baselines (the gap is less than 0.15dB). 
However, as a nonlinear representation method, MLP still has advantages over our linear representation with low scale factors. See Sec. \ref{sec:discuss1} for discussion on this issue.

\noindent{\bf Qualitative results.} 
Fig. \ref{results_img} provides qualitative results with SOTA methods by using different scale factor. We show competitive visual quality against others, more results are provided in supplementary material. From the local perspective, LIIF and LTE only generate smooth patches, while our OPE with max frequency 3 is enough to achieve similar visual quality. We also notice LIIF \cite{liif} has artifact (vertical stripes) in the 1st row, this is a common drawback for implicit neural representation and is hard to be explained. However, with our image representation, there is no artifacts. In the 2nd row, we could observe a sprout (in red rectangle) in the GT, the same region of LIIF is vanished, and the boundary of our sprout is more obvious than LTE.

\noindent{\bf Computing efficiency of upsampling module.} 
We measure computing efficiency with MACs (multiply-accumulate operations), FLOPs (floating point operations) and actual running time. In Tab. \ref{tab:timemem} column 2-3, judged by the time complexity measured by the number of operations, we save 2 orders of magnitude. In our upsampling module, there is only one matrix operation and essential position encoding between input and output.
In Tab. \ref{div2k_val} we show shortest inference time benefiting from our compact SR framework. To further demonstrate our time advantage on large size images, we take $256\times 256$ as LR input of encoder and calculate time consumption of upsampling module with scale factor $\times 4$-$\times 30$ on NVIDIA RTX 3090. As shown in Tab. \ref{tab:scaletime} , our upsampling module shows 26\%-57\% time advantage, this advantage keeps growing with larger scale factor. Notice We do not take advantage of GPU acceleration to design the upsampling module carefully, with hardware optimization, we believe our time advantage could be much larger thanks to fewer number of operations required.

\noindent{\bf Memory consumption of upsampling module.} 
In Tab. \ref{tab:timemem} column 4-5 we compare GPU memory consumption of OPE-Upscale Module with LIIF \cite{liif} and LTE \cite{lee2022local} under training mode and testing mode of Pytorch \cite{pytorch}. For training mode, we use a $48\times 48$ patch as input and sample 2304 pixels as output following the default training strategy in arbitrary-scale SR works. For testing mode, we use $512 \times 512$ image as input with scale factor 4 (2K target image). As a interpretable image representation without network parameters, OPE-Upscale Module saves memory of intermediate data (e.g. gradients, hidden layer outputs), and this advantage is fully reflected in training mode.

\noindent{\bf Flipping consistency.} 
As described in Sec. \ref{sec:intro}, the INR-based upsampling module like \cite{liif} is sensitive for the flipping of feature map. However, our method solves this problem completely and elegantly. The orthogonal basis of OPE is based on symmetric sinusoidal function, which leads to advantage of our method for keeping the flipping consistency. Also, more samples are provided in supplementary material for verifying other more flipping transforms.

\begin{table}[t]
\centering
\renewcommand\arraystretch{1.2}
\setlength{\tabcolsep}{5.0pt}
\scriptsize
\begin{tabular}{c|c|c|c|c|c} 
Method & Params & MACs & FLOPs & Mem (training) & Mem (Test)\\
\hline
LIIF& 0.35 M &  429 K &  6.2 G     & 85.1 + 1.9 M           & 32 + 96 M \\
LTE&  0.26 M &  526 K & 7.5 G  & 97.8 + 1.9 M         & 64 + 96 M \\
OPE (ours)&  \textbf{0 M} &  \textbf{6 K} & \textbf{85 M} & \textbf{0 + 1.9 M} & \textbf{0 + 96 M} \\
\end{tabular}

\vspace{-6pt}
\caption{\textbf{Parameter number, time complexity and memory consumption.} MACs: multiply-accumulate operations, FLOPs: floating point operations, Mem: intermediate data + essential output for GPU memory consumption. We use $n=3$ as maximum frequency of OPE and test in training mode and test mode on Pytorch with tool: torch.cuda.memory\_allocated(). Training mode: $48^2$ to 2304 pixels, test mode: $512^2$ to $2048^2$.}

\label{tab:timemem}
\vspace{-6pt}
\end{table}
\begin{table}[t]
\centering
\renewcommand\arraystretch{1.2}
\setlength{\tabcolsep}{5.0pt}
\scriptsize
\begin{tabular}{c|c|c|c|c|c|c|c} 
Method & $\times 4$ & $\times 8$ & $\times 12$ & $\times 16$ & $\times 20$& $\times 24$& $\times 30$\\
\hline
LIIF& 382 &  1521 &  3530   & 6004 & 10274 & 18350 & 27866\\
LTE& 376 &  1490  & 3340    & 5922 & 10268 & 18340 & 27838\\
OPE (ours)&  \textbf{277} &  \textbf{1125} & \textbf{2495} & \textbf{3719}& \textbf{5673} & \textbf{8366}& \textbf{12012} \\
\hline
percentage &  28\% & 26\% & 30\% & 39\%& 45\%& 55\% & 57\% \\
\end{tabular}

 \caption{\textbf{Rendering time of upsampling module (ms/Img)} with input size $256\times 256$. Last line: time saving percentage. We use $n=3$ as maximum frequency of OPE. Our time advantage grows as the rendering resolution increases. We save 40\% rendering time in average.}
\label{tab:scaletime}
\vspace*{-6pt}
\end{table}

\begin{table}[t]
\centering
\renewcommand\arraystretch{1.2}
\setlength{\tabcolsep}{5.0pt}
\scriptsize
\begin{tabular}{c
|>{\centering\arraybackslash}p{0.80cm}>{\centering\arraybackslash}p{0.80cm}>{\centering\arraybackslash}p{0.80cm}
|>{\centering\arraybackslash}p{0.80cm}>{\centering\arraybackslash}p{0.80cm}
}
& \multicolumn{3}{c|}{In-scale} & \multicolumn{2}{c}{Out-scale} \\
& $\times 2$ & $\times 3$ & $\times 4$ & $\times 6$ & $\times 8$\\
\hline

OPE & \textbf{33.29} & \textbf{30.29} & \textbf{28.65} & \textbf{26.46} & \textbf{24.98}\\
OPE (-E)  & 33.27 & 30.23 & 28.56 & 26.34  & 24.82 \\
OPE (-I)  & 33.28 & 30.26 & 28.63 & 26.44  & 24.97 \\
OPE (-IE) & 33.20 & 30.09 & 28.44 & 26.25  & 24.70 \\
\end{tabular}
\vspace*{-6pt}
\caption{\textbf{Quantitative ablation study of OPE on Set14.} EDSR-baseline \cite{edsr} is used as encoder.}
\label{tab: ablation}
\vspace{-15pt}
\end{table}

\begin{figure}[t]
\footnotesize
\centering
\stackunder[2pt]{\includegraphics[width=0.8in]{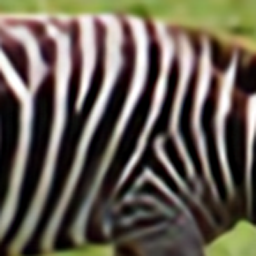}}{OPE}\hspace{-1pt}
\stackunder[2pt]{\includegraphics[width=0.8in]{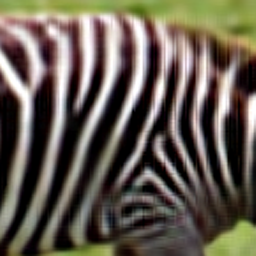}}{OPE(-E)}\hspace{-1pt}
\stackunder[2pt]{\includegraphics[width=0.8in]{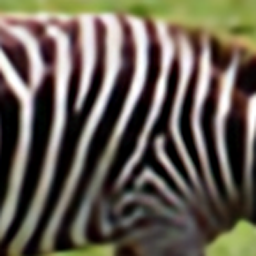}}{OPE(-I)}\hspace{-1pt}
\stackunder[2pt]{\includegraphics[width=0.8in]{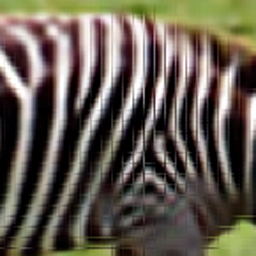}}{OPE(-IE)}\hspace{-1pt}

\vspace{-6pt}
\caption{\textbf{Qualitative ablation study about patch ensemble on Set14.} I: local ensemble styled interpolation, E: extension of relative coordinate domain. \textbf{OPE(-E)}: OPE-Upscale module without E but with I. \textbf{OPE(-I)}: OPE-Upscale module without I but with E. \textbf{OPE(-IE)}: OPE-Upscale module without patch ensemble.}

\label{fig: ablation}
\end{figure}
\vspace{-6pt}

\subsection{Ablation Study}
In this section, we show the effect of local ensemble styled interpolation (I) Eq.(\ref{eq9}) and extension of relative coordinate domain (E) Eq.(\ref{eq10}) in patch ensemble. We choose EDSR-baseline \cite{edsr} as encoder and compare four settings. (OPE): OPE-Upscale module with I and E. (OPE-E): OPE-Upscale module without E but with I. (OPE-I): OPE-Upscale module without I but with E. (OPE-IE): OPE-Upscale module without I and E (that is, without patch ensemble).

Fig. \ref{fig: ablation} and Tab. \ref{tab: ablation} show the comparison. The (OPE-E) is unable to rescale the relative coordinate domain of other three further latent code to $[-1,1]\times[-1,1]$ , hence brings periodic stripes on the SR image. 
The (OPE-I) result has no obvious discontinuity between patches, since E plays a positive role, however, this means only small region of patch is presented in target image. The (OPE-IE) result in obvious dense boundary between patches, which proves that E and I are both indispensable. 
\section{Discussions}
\label{sec:discuss}
\noindent{\bf Low scale factor.} 
\label{sec:discuss1}
In Tab. \ref{tab:Quan_Bench} and Tab. \ref{div2k_val}, we observe our quantitative results decreasing in low scale factor, especially when input size is small like benchmark datasets. 
Since small target image size ($W\times H$) means a larger grid in the 2D domain ($[-1, 1]\times [-1, 1]$), only utilizing one central point value to represent the entire larger grid would lose detailed information than a smaller grid.
Since 2D domain we applied is continuous, the higher the resolution of target image is, the stronger representation ability we could achieve.
For MLP-based representation \cite{liif,lee2022local}, the nonlinear operation could well avoid this. In our method, this defect can be ignored for high SR scale factors where pixels are dense, while for low scale as $\times 2,\times 3$, our performance just degrades slightly. This issue could be solved by sampling more points for every grid region and calculate their mean value with careful time consumption trade-off, this is a direction worthy exploring.


\section{Conclusion}
\label{sec:conclusion}
In this paper, we propose an interpretable method for continuous image representation without implicit neural network. 
We propose a novel position encoding method, which is in form of 2D-Fourier Series and naturally corresponds to 2D image coordinates. Our OPE is theoretically proved as a set of orthogonal basis in inner product space, which is both interpretable and rich in representation. Based on OPE, we further propose OPE-Upscale Module, which is parameter-free for arbitrary-scale image super-resolution. OPE-Upscale Module simplifies the existing deep SR framework, leads to high computing efficiency and less memory consumption. 
Our OPE-Upscale Module could be easily integrated into existing image super-resolution pipeline, and the extensive experiments prove that our method has competitive results with SOTA. In addition, we also provide an explanation of currently used sinusoidal position encoding from the perspective of orthogonal basis, for the future direction, more position encoding could be explored. (e.g. "Legendre position encoding" or "Chebyshev position encoding" may also works for MLP.)


{\small
\bibliographystyle{ieee_fullname}
\bibliography{egbib}

\begin{thebibliography}{10}\itemsep=-1pt

\bibitem{div2k}
Eirikur Agustsson and Radu Timofte.
\newblock Ntire 2017 challenge on single image super-resolution: Dataset and
  study.
\newblock In {\em Proceedings of the IEEE conference on computer vision and
  pattern recognition workshops}, pages 126--135, 2017.

\bibitem{sh1}
Ronen Basri and David~W Jacobs.
\newblock Lambertian reflectance and linear subspaces.
\newblock {\em IEEE transactions on pattern analysis and machine intelligence},
  25(2):218--233, 2003.

\bibitem{set5}
Marco Bevilacqua, Aline Roumy, Christine Guillemot, and Marie~Line
  Alberi-Morel.
\newblock Low-complexity single-image super-resolution based on nonnegative
  neighbor embedding.
\newblock 2012.

\bibitem{liif}
Yinbo Chen, Sifei Liu, and Xiaolong Wang.
\newblock Learning continuous image representation with local implicit image
  function.
\newblock In {\em Proceedings of the IEEE/CVF conference on computer vision and
  pattern recognition}, pages 8628--8638, 2021.

\bibitem{choi2021toward}
Jooyoung Choi, Jungbeom Lee, Yonghyun Jeong, and Sungroh Yoon.
\newblock Toward spatially unbiased generative models.
\newblock {\em arXiv preprint arXiv:2108.01285}, 2021.

\bibitem{daubechies1992ten}
Ingrid Daubechies.
\newblock {\em Ten lectures on wavelets}.
\newblock SIAM, 1992.

\bibitem{srcnn}
Chao Dong, Chen~Change Loy, Kaiming He, and Xiaoou Tang.
\newblock Learning a deep convolutional network for image super-resolution.
\newblock In {\em European conference on computer vision}, pages 184--199.
  Springer, 2014.

\bibitem{accelerating}
Chao Dong, Chen~Change Loy, and Xiaoou Tang.
\newblock Accelerating the super-resolution convolutional neural network.
\newblock In {\em European conference on computer vision}, pages 391--407.
  Springer, 2016.

\bibitem{lras}
Vincent Dumoulin, Jonathon Shlens, and Manjunath Kudlur.
\newblock A learned representation for artistic style.
\newblock {\em arXiv preprint arXiv:1610.07629}, 2016.

\bibitem{plenoxels}
Sara Fridovich-Keil, Alex Yu, Matthew Tancik, Qinhong Chen, Benjamin Recht, and
  Angjoo Kanazawa.
\newblock Plenoxels: Radiance fields without neural networks.
\newblock In {\em Proceedings of the IEEE/CVF Conference on Computer Vision and
  Pattern Recognition}, pages 5501--5510, 2022.

\bibitem{local}
Kyle Genova, Forrester Cole, Avneesh Sud, Aaron Sarna, and Thomas Funkhouser.
\newblock Local deep implicit functions for 3d shape.
\newblock In {\em Proceedings of the IEEE/CVF Conference on Computer Vision and
  Pattern Recognition}, pages 4857--4866, 2020.

\bibitem{dbp}
Muhammad Haris, Gregory Shakhnarovich, and Norimichi Ukita.
\newblock Deep back-projection networks for super-resolution.
\newblock In {\em Proceedings of the IEEE conference on computer vision and
  pattern recognition}, pages 1664--1673, 2018.

\bibitem{universal}
Kurt Hornik, Maxwell Stinchcombe, and Halbert White.
\newblock Multilayer feedforward networks are universal approximators.
\newblock {\em Neural networks}, 2(5):359--366, 1989.

\bibitem{metasr}
Xuecai Hu, Haoyuan Mu, Xiangyu Zhang, Zilei Wang, Tieniu Tan, and Jian Sun.
\newblock Meta-sr: A magnification-arbitrary network for super-resolution.
\newblock In {\em Proceedings of the IEEE/CVF conference on computer vision and
  pattern recognition}, pages 1575--1584, 2019.

\bibitem{pref}
Binbin Huang, Xinhao Yan, Anpei Chen, Shenghua Gao, and Jingyi Yu.
\newblock Pref: Phasorial embedding fields for compact neural representations.
\newblock {\em arXiv preprint arXiv:2205.13524}, 2022.

\bibitem{urban100}
Jia-Bin Huang, Abhishek Singh, and Narendra Ahuja.
\newblock Single image super-resolution from transformed self-exemplars.
\newblock In {\em Proceedings of the IEEE conference on computer vision and
  pattern recognition}, pages 5197--5206, 2015.

\bibitem{karras2021alias}
Tero Karras, Miika Aittala, Samuli Laine, Erik H{\"a}rk{\"o}nen, Janne
  Hellsten, Jaakko Lehtinen, and Timo Aila.
\newblock Alias-free generative adversarial networks.
\newblock {\em Advances in Neural Information Processing Systems}, 34:852--863,
  2021.

\bibitem{accurate}
Jiwon Kim, Jung~Kwon Lee, and Kyoung~Mu Lee.
\newblock Accurate image super-resolution using very deep convolutional
  networks.
\newblock In {\em Proceedings of the IEEE conference on computer vision and
  pattern recognition}, pages 1646--1654, 2016.

\bibitem{dr}
Jiwon Kim, Jung~Kwon Lee, and Kyoung~Mu Lee.
\newblock Deeply-recursive convolutional network for image super-resolution.
\newblock In {\em Proceedings of the IEEE conference on computer vision and
  pattern recognition}, pages 1637--1645, 2016.

\bibitem{adam}
Diederik~P Kingma and Jimmy Ba.
\newblock Adam: A method for stochastic optimization.
\newblock {\em arXiv preprint arXiv:1412.6980}, 2014.

\bibitem{poly1}
Tom Koornwinder.
\newblock Two-variable analogues of the classical orthogonal polynomials.
\newblock In {\em Theory and application of special functions}, pages 435--495.
  Elsevier, 1975.

\bibitem{lap}
Wei-Sheng Lai, Jia-Bin Huang, Narendra Ahuja, and Ming-Hsuan Yang.
\newblock Deep laplacian pyramid networks for fast and accurate
  super-resolution.
\newblock In {\em Proceedings of the IEEE conference on computer vision and
  pattern recognition}, pages 624--632, 2017.

\bibitem{deeplap}
Wei-Sheng Lai, Jia-Bin Huang, Narendra Ahuja, and Ming-Hsuan Yang.
\newblock Fast and accurate image super-resolution with deep laplacian pyramid
  networks.
\newblock {\em IEEE transactions on pattern analysis and machine intelligence},
  41(11):2599--2613, 2018.

\bibitem{pattern2}
Seyed~Mehdi Lajevardi and Zahir~M Hussain.
\newblock Higher order orthogonal moments for invariant facial expression
  recognition.
\newblock {\em Digital Signal Processing}, 20(6):1771--1779, 2010.

\bibitem{srgan}
Christian Ledig, Lucas Theis, Ferenc Husz{\'a}r, Jose Caballero, Andrew
  Cunningham, Alejandro Acosta, Andrew Aitken, Alykhan Tejani, Johannes Totz,
  Zehan Wang, et~al.
\newblock Photo-realistic single image super-resolution using a generative
  adversarial network.
\newblock In {\em Proceedings of the IEEE conference on computer vision and
  pattern recognition}, pages 4681--4690, 2017.

\bibitem{lee2022local}
Jaewon Lee and Kyong~Hwan Jin.
\newblock Local texture estimator for implicit representation function.
\newblock In {\em Proceedings of the IEEE/CVF Conference on Computer Vision and
  Pattern Recognition}, pages 1929--1938, 2022.

\bibitem{feedback}
Zhen Li, Jinglei Yang, Zheng Liu, Xiaomin Yang, Gwanggil Jeon, and Wei Wu.
\newblock Feedback network for image super-resolution.
\newblock In {\em Proceedings of the IEEE/CVF conference on computer vision and
  pattern recognition}, pages 3867--3876, 2019.

\bibitem{swinir}
Jingyun Liang, Jiezhang Cao, Guolei Sun, Kai Zhang, Luc Van~Gool, and Radu
  Timofte.
\newblock Swinir: Image restoration using swin transformer.
\newblock In {\em Proceedings of the IEEE/CVF International Conference on
  Computer Vision}, pages 1833--1844, 2021.

\bibitem{edsr}
Bee Lim, Sanghyun Son, Heewon Kim, Seungjun Nah, and Kyoung Mu~Lee.
\newblock Enhanced deep residual networks for single image super-resolution.
\newblock In {\em Proceedings of the IEEE conference on computer vision and
  pattern recognition workshops}, pages 136--144, 2017.

\bibitem{lin2021infinitygan}
Chieh~Hubert Lin, Hsin-Ying Lee, Yen-Chi Cheng, Sergey Tulyakov, and Ming-Hsuan
  Yang.
\newblock Infinitygan: Towards infinite-pixel image synthesis.
\newblock {\em arXiv preprint arXiv:2104.03963}, 2021.

\bibitem{liu2020neural}
Lingjie Liu, Jiatao Gu, Kyaw Zaw~Lin, Tat-Seng Chua, and Christian Theobalt.
\newblock Neural sparse voxel fields.
\newblock {\em Advances in Neural Information Processing Systems},
  33:15651--15663, 2020.

\bibitem{sparse2}
Julien Mairal, Michael Elad, and Guillermo Sapiro.
\newblock Sparse representation for color image restoration.
\newblock {\em IEEE Transactions on image processing}, 17(1):53--69, 2007.

\bibitem{mallat1999wavelet}
St{\'e}phane Mallat.
\newblock {\em A wavelet tour of signal processing}.
\newblock Elsevier, 1999.

\bibitem{b100}
David Martin, Charless Fowlkes, Doron Tal, and Jitendra Malik.
\newblock A database of human segmented natural images and its application to
  evaluating segmentation algorithms and measuring ecological statistics.
\newblock In {\em Proceedings Eighth IEEE International Conference on Computer
  Vision. ICCV 2001}, volume~2, pages 416--423. IEEE, 2001.

\bibitem{occupancy}
Lars Mescheder, Michael Oechsle, Michael Niemeyer, Sebastian Nowozin, and
  Andreas Geiger.
\newblock Occupancy networks: Learning 3d reconstruction in function space.
\newblock In {\em Proceedings of the IEEE/CVF conference on computer vision and
  pattern recognition}, pages 4460--4470, 2019.

\bibitem{nerf}
Ben Mildenhall, Pratul~P Srinivasan, Matthew Tancik, Jonathan~T Barron, Ravi
  Ramamoorthi, and Ren Ng.
\newblock Nerf: Representing scenes as neural radiance fields for view
  synthesis.
\newblock {\em Communications of the ACM}, 65(1):99--106, 2021.

\bibitem{niemeyer2021giraffe}
Michael Niemeyer and Andreas Geiger.
\newblock Giraffe: Representing scenes as compositional generative neural
  feature fields.
\newblock In {\em Proceedings of the IEEE/CVF Conference on Computer Vision and
  Pattern Recognition}, pages 11453--11464, 2021.

\bibitem{volume}
Michael Niemeyer, Lars Mescheder, Michael Oechsle, and Andreas Geiger.
\newblock Differentiable volumetric rendering: Learning implicit 3d
  representations without 3d supervision.
\newblock In {\em Proceedings of the IEEE/CVF Conference on Computer Vision and
  Pattern Recognition}, pages 3504--3515, 2020.

\bibitem{texf}
Michael Oechsle, Lars Mescheder, Michael Niemeyer, Thilo Strauss, and Andreas
  Geiger.
\newblock Texture fields: Learning texture representations in function space.
\newblock In {\em Proceedings of the IEEE/CVF International Conference on
  Computer Vision}, pages 4531--4540, 2019.

\bibitem{deepsdf}
Jeong~Joon Park, Peter Florence, Julian Straub, Richard Newcombe, and Steven
  Lovegrove.
\newblock Deepsdf: Learning continuous signed distance functions for shape
  representation.
\newblock In {\em Proceedings of the IEEE/CVF conference on computer vision and
  pattern recognition}, pages 165--174, 2019.

\bibitem{pytorch}
Adam Paszke, Sam Gross, Francisco Massa, Adam Lerer, James Bradbury, Gregory
  Chanan, Trevor Killeen, Zeming Lin, Natalia Gimelshein, Luca Antiga, et~al.
\newblock Pytorch: An imperative style, high-performance deep learning library.
\newblock {\em Advances in neural information processing systems}, 32, 2019.

\bibitem{convoccu}
Songyou Peng, Michael Niemeyer, Lars Mescheder, Marc Pollefeys, and Andreas
  Geiger.
\newblock Convolutional occupancy networks.
\newblock In {\em European Conference on Computer Vision}, pages 523--540.
  Springer, 2020.

\bibitem{spectral}
Nasim Rahaman, Aristide Baratin, Devansh Arpit, Felix Draxler, Min Lin, Fred
  Hamprecht, Yoshua Bengio, and Aaron Courville.
\newblock On the spectral bias of neural networks.
\newblock In {\em International Conference on Machine Learning}, pages
  5301--5310. PMLR, 2019.

\bibitem{sh2}
Ravi Ramamoorthi and Pat Hanrahan.
\newblock On the relationship between radiance and irradiance: determining the
  illumination from images of a convex lambertian object.
\newblock {\em JOSA A}, 18(10):2448--2459, 2001.

\bibitem{schwarz2020graf}
Katja Schwarz, Yiyi Liao, Michael Niemeyer, and Andreas Geiger.
\newblock Graf: Generative radiance fields for 3d-aware image synthesis.
\newblock {\em Advances in Neural Information Processing Systems},
  33:20154--20166, 2020.

\bibitem{subpixel}
Wenzhe Shi, Jose Caballero, Ferenc Husz{\'a}r, Johannes Totz, Andrew~P Aitken,
  Rob Bishop, Daniel Rueckert, and Zehan Wang.
\newblock Real-time single image and video super-resolution using an efficient
  sub-pixel convolutional neural network.
\newblock In {\em Proceedings of the IEEE conference on computer vision and
  pattern recognition}, pages 1874--1883, 2016.

\bibitem{zero}
Assaf Shocher, Nadav Cohen, and Michal Irani.
\newblock “zero-shot” super-resolution using deep internal learning.
\newblock In {\em Proceedings of the IEEE conference on computer vision and
  pattern recognition}, pages 3118--3126, 2018.

\bibitem{paf}
Vincent Sitzmann, Julien Martel, Alexander Bergman, David Lindell, and Gordon
  Wetzstein.
\newblock Implicit neural representations with periodic activation functions.
\newblock {\em Advances in Neural Information Processing Systems},
  33:7462--7473, 2020.

\bibitem{deepvoxels}
Vincent Sitzmann, Justus Thies, Felix Heide, Matthias Nie{\ss}ner, Gordon
  Wetzstein, and Michael Zollhofer.
\newblock Deepvoxels: Learning persistent 3d feature embeddings.
\newblock In {\em Proceedings of the IEEE/CVF Conference on Computer Vision and
  Pattern Recognition}, pages 2437--2446, 2019.

\bibitem{srn}
Vincent Sitzmann, Michael Zollh{\"o}fer, and Gordon Wetzstein.
\newblock Scene representation networks: Continuous 3d-structure-aware neural
  scene representations.
\newblock {\em Advances in Neural Information Processing Systems}, 32, 2019.

\bibitem{sh3}
Peter-Pike Sloan, Jan Kautz, and John Snyder.
\newblock Precomputed radiance transfer for real-time rendering in dynamic,
  low-frequency lighting environments.
\newblock In {\em Proceedings of the 29th annual conference on Computer
  graphics and interactive techniques}, pages 527--536, 2002.

\bibitem{son2021srwarp}
Sanghyun Son and Kyoung~Mu Lee.
\newblock Srwarp: Generalized image super-resolution under arbitrary
  transformation.
\newblock In {\em Proceedings of the IEEE/CVF conference on computer vision and
  pattern recognition}, pages 7782--7791, 2021.

\bibitem{rrn}
Ying Tai, Jian Yang, and Xiaoming Liu.
\newblock Image super-resolution via deep recursive residual network.
\newblock In {\em Proceedings of the IEEE conference on computer vision and
  pattern recognition}, pages 3147--3155, 2017.

\bibitem{memnet}
Ying Tai, Jian Yang, Xiaoming Liu, and Chunyan Xu.
\newblock Memnet: A persistent memory network for image restoration.
\newblock In {\em Proceedings of the IEEE international conference on computer
  vision}, pages 4539--4547, 2017.

\bibitem{ntire}
Radu Timofte, Eirikur Agustsson, Luc Van~Gool, Ming-Hsuan Yang, and Lei Zhang.
\newblock Ntire 2017 challenge on single image super-resolution: Methods and
  results.
\newblock In {\em Proceedings of the IEEE conference on computer vision and
  pattern recognition workshops}, pages 114--125, 2017.

\bibitem{dsc}
Tong Tong, Gen Li, Xiejie Liu, and Qinquan Gao.
\newblock Image super-resolution using dense skip connections.
\newblock In {\em Proceedings of the IEEE international conference on computer
  vision}, pages 4799--4807, 2017.

\bibitem{vaswani2017attention}
Ashish Vaswani, Noam Shazeer, Niki Parmar, Jakob Uszkoreit, Llion Jones,
  Aidan~N Gomez, {\L}ukasz Kaiser, and Illia Polosukhin.
\newblock Attention is all you need.
\newblock {\em Advances in neural information processing systems}, 30, 2017.

\bibitem{wang2021learning}
Longguang Wang, Yingqian Wang, Zaiping Lin, Jungang Yang, Wei An, and Yulan
  Guo.
\newblock Learning a single network for scale-arbitrary super-resolution.
\newblock In {\em Proceedings of the IEEE/CVF international conference on
  computer vision}, pages 4801--4810, 2021.

\bibitem{realesrgan}
Xintao Wang, Liangbin Xie, Chao Dong, and Ying Shan.
\newblock Real-esrgan: Training real-world blind super-resolution with pure
  synthetic data.
\newblock In {\em Proceedings of the IEEE/CVF International Conference on
  Computer Vision}, pages 1905--1914, 2021.

\bibitem{progress}
Yifan Wang, Federico Perazzi, Brian McWilliams, Alexander Sorkine-Hornung, Olga
  Sorkine-Hornung, and Christopher Schroers.
\newblock A fully progressive approach to single-image super-resolution.
\newblock In {\em Proceedings of the IEEE conference on computer vision and
  pattern recognition workshops}, pages 864--873, 2018.

\bibitem{sevenways}
Yifan Wang, Federico Perazzi, Brian McWilliams, Alexander Sorkine-Hornung, Olga
  Sorkine-Hornung, and Christopher Schroers.
\newblock A fully progressive approach to single-image super-resolution.
\newblock In {\em Proceedings of the IEEE conference on computer vision and
  pattern recognition workshops}, pages 864--873, 2018.

\bibitem{srsurvey}
Zhihao Wang, Jian Chen, and Steven~CH Hoi.
\newblock Deep learning for image super-resolution: A survey.
\newblock {\em IEEE transactions on pattern analysis and machine intelligence},
  43(10):3365--3387, 2020.

\bibitem{sparse3}
John Wright, Allen~Y Yang, Arvind Ganesh, S~Shankar Sastry, and Yi Ma.
\newblock Robust face recognition via sparse representation.
\newblock {\em IEEE transactions on pattern analysis and machine intelligence},
  31(2):210--227, 2008.

\bibitem{xu2021ultrasr}
Xingqian Xu, Zhangyang Wang, and Humphrey Shi.
\newblock Ultrasr: Spatial encoding is a missing key for implicit image
  function-based arbitrary-scale super-resolution.
\newblock {\em arXiv preprint arXiv:2103.12716}, 2021.

\bibitem{sparse1}
Jianchao Yang, John Wright, Thomas Huang, and Yi Ma.
\newblock Image super-resolution as sparse representation of raw image patches.
\newblock In {\em 2008 IEEE conference on computer vision and pattern
  recognition}, pages 1--8. IEEE, 2008.

\bibitem{yoon2022spheresr}
Youngho Yoon, Inchul Chung, Lin Wang, and Kuk-Jin Yoon.
\newblock Spheresr: 360deg image super-resolution with arbitrary projection via
  continuous spherical image representation.
\newblock In {\em Proceedings of the IEEE/CVF Conference on Computer Vision and
  Pattern Recognition}, pages 5677--5686, 2022.

\bibitem{set14}
Roman Zeyde, Michael Elad, and Matan Protter.
\newblock On single image scale-up using sparse-representations.
\newblock In {\em International conference on curves and surfaces}, pages
  711--730. Springer, 2010.

\bibitem{pattern1}
Hui Zhang, Huazhong Shu, Guoniu~N Han, Gouenou Coatrieux, Limin Luo, and
  Jean~Louis Coatrieux.
\newblock Blurred image recognition by legendre moment invariants.
\newblock {\em IEEE Transactions on Image Processing}, 19(3):596--611, 2009.

\bibitem{zhang2020nerf++}
Kai Zhang, Gernot Riegler, Noah Snavely, and Vladlen Koltun.
\newblock Nerf++: Analyzing and improving neural radiance fields.
\newblock {\em arXiv preprint arXiv:2010.07492}, 2020.

\bibitem{rcan}
Yulun Zhang, Kunpeng Li, Kai Li, Lichen Wang, Bineng Zhong, and Yun Fu.
\newblock Image super-resolution using very deep residual channel attention
  networks.
\newblock In {\em Proceedings of the European conference on computer vision
  (ECCV)}, pages 286--301, 2018.

\bibitem{rdn}
Yulun Zhang, Yapeng Tian, Yu Kong, Bineng Zhong, and Yun Fu.
\newblock Residual dense network for image super-resolution.
\newblock In {\em Proceedings of the IEEE conference on computer vision and
  pattern recognition}, pages 2472--2481, 2018.

\bibitem{poly2}
Hongqing Zhu.
\newblock Image representation using separable two-dimensional continuous and
  discrete orthogonal moments.
\newblock {\em Pattern Recognition}, 45(4):1540--1558, 2012.

\end{thebibliography}
}

\end{document}